\documentclass[twoside]{article}

\usepackage[accepted]{aistats2024}
\usepackage[round]{natbib}

\usepackage[inline]{enumitem}
\usepackage{graphicx}
\usepackage{float}
\usepackage{hyperref}
\hypersetup{
    colorlinks=false,
    linkcolor=blue,
    filecolor=magenta,      
    urlcolor=cyan,
    pdftitle={Overleaf Example},
    pdfpagemode=FullScreen,
    }
\usepackage{lineno,amsthm}

\usepackage{algorithm}
\usepackage{algpseudocode}

\algnewcommand\INPUT{\item[\textbf{Input:}]}%
\algnewcommand\OUTPUT{\item[\textbf{Output:}]}%

\newcommand{\bd}{\boldsymbol}

\newcommand{\be}{\begin{equation}}
\newcommand{\ee}{\end{equation}}

\newtheorem{theorem}{Theorem}

\begin{document}

%

%

\twocolumn[

\aistatstitle{DeepFDR: A Deep Learning-based False Discovery Rate Control Method for Neuroimaging Data}

\aistatsauthor{Taehyo Kim$^{\dag,1}$ \And Hai Shu$^{\dag,\ddag,1}$ \And  Qiran Jia$^{1,2}$ \And Mony J. de Leon$^{3}$}

\centering{for the Alzheimer’s Disease Neuroimaging Initiative\ref{datause}
}

\aistatsaddress{ $^1$Department of Biostatistics, School of Global Public Health, New York University } 
\vspace{-0.8cm}
\aistatsaddress{ 
$^2$Department of Population and Public Health Sciences, University of Southern California 
}
\vspace{-0.8cm}
\aistatsaddress{ 
  $^3$Brain Health
Imaging Institute, Department of Radiology,
Weill Cornell Medicine 
}
\vspace{-0.8cm}
\aistatsaddress{ 
  $^\dag$Equal contributions
}
\vspace{-0.8cm}
\aistatsaddress{ 
  $^\ddag$Correspondence: hs120@nyu.edu
}

]

\begin{abstract}
Voxel-based multiple testing is widely used in  neuroimaging data analysis. Traditional false discovery rate (FDR) control methods often ignore the spatial dependence among the voxel-based tests and thus suffer from substantial loss of testing power. While recent spatial FDR control methods have emerged, their validity and optimality 
remain questionable when handling 
the complex spatial dependencies of the brain.
Concurrently, deep learning methods have revolutionized image segmentation, a task closely related to voxel-based multiple testing.
In this paper, we propose DeepFDR, a novel spatial FDR control method that
leverages unsupervised deep learning-based image segmentation  to address the voxel-based multiple testing problem. 
Numerical studies, including comprehensive simulations and Alzheimer's disease FDG-PET image analysis, demonstrate DeepFDR's superiority over existing methods.
DeepFDR not only  excels in FDR control and effectively diminishes the false nondiscovery rate, but also boasts exceptional computational efficiency highly suited for tackling large-scale neuroimaging data.

\end{abstract}

\vspace{-0.3cm}
\section{INTRODUCTION}\label{sec: intro}
\vspace{-0.3cm}

Voxel-based multiple testing is widely used in neuroimaging data analysis~\citep{Ashb00,Geno02, mirman2018corrections}. For instance, in Alzheimer's disease  research, as a neurodegeneration biomarker, 
Fluorine-18 fluorodeoxyglucose positron emission tomography (FDG-PET) measures the 
brain glucose metabolism and is extensively used for early diagnosis and monitoring the progression of Alzheimer's disease~\citep{Alex02, Drze03, Shiv15, Ou19}. To statistically compare brain glucose metabolism between two groups of different disease statuses, FDG-PET studies in Alzheimer's disease often conduct multiple testing at the voxel level to identify brain regions with functional abnormalities~\citep{Mosc05,Lee5,Shu15,kantarci2021fdg}.

The prevalent multiple testing methods are based on controlling the {\it false discovery rate} (FDR; \citet{Benj95}), an alternative yet more powerful measure of type I error than the conventional family-wise error rate (FWER).
The corresponding measure of type II error is the {\it false nondiscovery rate} (FNR; \cite{Geno02b}).
 However, for neuroimaging data, traditional FDR control methods 
 such as the BH~\citep{Benj95},
 q-value~\citep{Stor03}, and LocalFDR~\citep{efron2004large} methods, ignore the spatial dependence among the voxel-based tests  and thus suffer from substantial loss of testing power~\citep{Shu15}. 
The voxel-based tests are inherently dependent due to the spatial structure among  brain voxels. 
 Although 
 some FDR control methods, 
 applicable to spatial and three-dimensional (3D) contexts,
 recently have been developed,
 they either use basic spatial models such as simple hidden Markov random fields (HMRF; \cite{Shu15,10.1214/16-AOAS956,kim2018peeling}) and simple Gaussian random fields \citep{Sun15}, or rely on local smoothing approaches \citep{Tans18,laws,han2023spatially}.
The {\it validity} of these methods in controlling  FDR and their {\it optimality} in minimizing FNR are called into question when handling the imaging data of the complex human brain, which is spatially heterogeneous due to its anatomical structure~\citep{Brod07} and exhibits long-distance functional connectivity between brain regions~\citep{Liu13}. 
Hence, it is imperative to develop a spatial FDR control method that  effectively  captures the brain's intricate dependencies and enjoys theoretical guarantees of the validity and optimality.

It is noteworthy that the aforementioned 
methods of \citet{Shu15}, \citet{10.1214/16-AOAS956}, \citet{kim2018peeling} and \citet{Sun15}
all use a testing procedure 
introduced by \citet{Sun09}, which relies  on the local index of significance (LIS)
rather than the more commonly used p-value.
Unlike the p-value, which is determined solely by 
the test statistic at the corresponding spatial location, the LIS at any spatial location is the conditional probability that its null hypothesis is true, given the test statistics from all spatial locations.
Under mild conditions,
the LIS-based testing procedure can asymptotically minimize the FNR while controlling the FDR under a prespecified level~\citep{Sun09,xie2011optimal}. 
Thus, the performance of the LIS-based 
testing procedure
hinges on the capability of the selected spatial model to appropriately model the dependencies among the tests.


A task closely related  to voxel-based multiple testing is image segmentation~\citep{minaee2021image}. Both  follow a procedure where the input is an image: a map of test statistics for multiple testing, and the target image for segmentation; the output assigns  a label to each voxel/pixel:
hypothesis state labels
in multiple testing, and segmentation labels in image segmentation.
This similarity  prompts the question: can we apply image segmentation models to voxel-based multiple testing?

In medical image segmentation, 
 deep learning methods,
especially the U-net and its variants~\citep{Ronn15,Ccic16,isensee2021nnu,chen2021transunet, cao2022swin,hatamizadeh2022unetr,pan2023eg},
have established state-of-the-art results.
The foundational U-net architecture  
consists of  a contracting path
designed to extract the global salient features and an expanding
path utilized to recover local spatial details through skip connections from the contracting path.
This innovative network design
empowers these network models to effectively capture both short and long-range spatial dependencies and account for spatial heterogeneity.
The U-net and its variants
have emerged as top performers in various segmentation tasks for neuroimaging data.
These include challenges like
the Brain Tumor Segmentation (BraTS) Challenge~\citep{bakas2018identifying},
the Ischemic Stroke Lesion Segmentation (ISLES) Challenge~\citep{liew2022large},
and the Infant Brain MRI Segmentation (iSeg) Challenge~\citep{sun2021multi}.

However, our voxel-based multiple testing is an unsupervised
learning task without ground-truth hypothesis state labels,
contrasting with
most deep-learning methods for image segmentation, which are supervised  and
require predefined ground-truth
labels during training~\citep{siddique2021u}. 
Recently, several unsupervised deep learning-based image segmentation methods have been developed.
\citet{xia2017wnet} proposed the W-net, a cascade of two U-nets, 
where the normalized
cut loss of the first U-net
and the reconstruction loss
of the second U-net 
are iteratively minimized to
generate segmentation probability maps.
\citet{kanezaki2018unsupervised}
utilized a convolutional neural network (CNN) to extract features, clustered them for pseudo labels, and alternately optimized the pseudo labels and segmentation network through self-training.
\citet{kim2020unsupervised} further improved upon this approach by introducing a spatial continuity loss.
\citet{pu2023deep} designed 
an autoencoder network integrated with an expectation-maximization module, which employs
a Gaussian mixture model 
to relate 
segmentation labels
to the deep features extracted from 
the encoder and constrained by
image reconstruction via the decoder,
and ultimately assigns labels based on their conditional probabilities given these deep features.

In this paper, we propose DeepFDR,
a novel deep learning-based FDR control method
for voxel-based multiple testing.
We innovatively connect
the voxel-based multiple testing
with the deep learning-based
unsupervised image segmentation.
Specifically,
we adopt the LIS-based testing procedure~\citep{Sun09},
where the LIS values
are estimated by the segmentation probability maps from
our modified version of the W-net~\citep{xia2017wnet}.
The aforementioned unsupervised
image segmentation methods
of \citet{kanezaki2018unsupervised},
\citet{kim2020unsupervised} and \citet{pu2023deep} are not applicable  in this context, as they
do not estimate the conditional probability of each voxel's label given the input image, which coincides with the LIS when the input is the map of test statistics.

To the best of our knowledge,
our work is the first to directly apply deep learning
to unsupervised spatial multiple testing. 
We notice that four recent studies \citep{xia2017neuralfdr,tansey2018black, romano2020deep, marandon2022machine} have also used 
deep neural networks in multiple testing,
but there are intrinsic distinctions between their approaches and ours.
\cite{xia2017neuralfdr} proposed the NeuralFDR method
to address multiple testing problems when covariate information for each hypothesis test is available.
NeuralFDR employs a deep neural network
to learn the p-value threshold as a function of
the covariates. 
Although 3D coordinates may serve as covariates, 
NeuralFDR assumes that  
the p-value and covariates for each test 
are independent under the null hypothesis but dependent under the alternative. This assumption does not align with the nature of spatial data, where true and false nulls can be spatially adjacent.
\cite{tansey2018black} developed the BB-FDR method for independent  tests each with covariates, in contrast to the dependent tests in our study. 
BB-FDR uses a deep neural network
to model the hyperprior parameters of the hypothesis state based on the covariates.
\cite{romano2020deep} introduced
Deep Knockoffs, a method that employs a deep neural network to generate model-X
knockoffs,
but
their model-X knockoffs problem is different from our voxel-based multiple testing problem.
\cite{marandon2022machine}
applied neural networks as classifiers to solve a semi-supervised multiple testing problem, where a subset of the sample data, termed a null training sample (NTS), is known from the null distribution. 
Their method is not applicable to our unsupervised voxel-based multiple testing due to the absence of an NTS.
In our context, even if an NTS might be additionally generated from a known null distribution, it would not offer useful spatial dependence information.

Our contributions are summarized as follows:
\vspace{-0.2cm}
\begin{itemize}[leftmargin=*, itemsep=0pt, topsep=0pt, partopsep=0pt, parsep=0pt]
\item We propose DeepFDR, a pioneering method that harmoniously combines deep learning techniques with voxel-based multiple testing. Inspired by advancements in unsupervised image segmentation, DeepFDR offers a fresh perspective on controlling the FDR in neuroimaging analyses.

\item We empirically demonstrate the superior performance of DeepFDR through rigorous simulation studies and in-depth analysis of 3D FDG-PET images pertaining to Alzheimer's disease. Our findings indicate its consistent capability to adeptly control FDR whilst effectively reducing FNR, thereby ensuring enhanced reliability of results.

\item DeepFDR exhibits exceptional  computational efficiency by leveraging the mature software and advanced optimization algorithms from deep learning.
This advantage distinguishes it from existing spatial FDR control methods, rendering it highly suited for handling large-scale neuroimaging data.

\end{itemize}

A Python package for our  DeepFDR
method is available at \url{https://github.com/kimtae55/DeepFDR}.

\section{METHOD}
\vspace{-0.3cm}
\subsection{Problem Formulation}
\vspace{-0.3cm}
Consider two population groups, for example, the Alzheimer's disease group and the cognitively normal group. We aim to compare 
the brain glucose metabolism
between the two groups by
testing the difference in their voxel-level
population means of the standardized uptake value ratio (SUVR) from FDG-PET.
Each subject
in the sample data
has a 3D brain FDG-PET image with $m$ voxels of interest.
Let $x_i$ be a test statistic for the null hypothesis $\mathcal{H}_i$, which assumes that there is no difference in the mean values of SUVR between the two groups at voxel~$i$.
The unobservable state label
$h_i$ is defined as $h_i = 1$ if $\mathcal{H}_i$ is false and $h_i = 0$ otherwise. The goal of  multiple testing is to predict the unknown labels $\bd{h} = [h_1, ..., h_m]$ based on the test statistics $\bd{x} = [x_1, ...,x_m]$. 
Table~\ref{table:c_matrix} summarizes the classification of tested hypotheses.
The FDR and FNR are defined as 
\be
FDR = E\left[\frac{N_{10}}{R\vee 1}\right]~~~\text{and}~~~
FNR = E\left[\frac{N_{01}}{A\vee 1}\right],
\ee
where $a\vee b=\max(a,b)$.
An FDR control method is {\it valid}
if it controls FDR at a prespecified level,
and is {\it optimal} if it has the smallest FNR among all valid FDR control methods.
We aim to develop an optimal FDR control method for voxel-based multiple testing.
For simplicity, false nulls
and rejected nulls are called {\it signals} and {\it discoveries}, respectively.

\begin{table}[h]
\begin{center}
\begin{tabular}{c|cccc}
\hline
Number	& Not rejected & Rejected & Total \\
\hline
True null & $N_{00}$ & $N_{10}$ & $m_0$\\
False null & $N_{01}$ & $N_{11}$ & $m_1$\\
Total & $A$ & $R$ &$m$\\
\hline
\end{tabular}\vspace{-0.3cm}
\caption{Classification of tested hypotheses} \label{table:c_matrix}
\end{center}\vspace{-0.3cm}
\end{table}

\vspace{-0.3cm}
\subsection{LIS-based Testing Procedure}
\vspace{-0.3cm}
\cite{Sun09} defined the LIS for hypothesis $\mathcal{H}_i$ by 
\be\label{def: LIS value}
LIS_i(\bd{x}) = P(h_i = 0|\bd{x}), 
\ee
which depends on all test statistics $\bd{x} = [x_1, ...,x_m]$, not just the local statistic $x_i$.
They proposed the LIS-based testing procedure
for controlling FDR at 
a prespecified level~$\alpha$:
\vspace{-0.3cm}
\be
\begin{split}\label{def: LIS procedure}
&\mbox{Let } k = \max\{j: \frac{1}{j}\sum_{i=1}^{j}LIS_{(i)}(\bd{x}) \leq \alpha\}, \\
&\mbox{then reject all } \mathcal{H}_{(i)} 
\mbox{ with } i = 1,...,k.
\end{split}
\ee
Here, $LIS_{(1)}(\bd{x}),\dots, LIS_{(m)}(\bd{x})$ are the ranked LIS values in ascending order and 
$\mathcal{H}_{(1)},\dots,\mathcal{H}_{(m)}$
are the corresponding null hypotheses.
In this procedure, $\{LIS_{i}(\bd{x})\}_{i=1}^m$ are practically  replaced with their estimates, denoted by $\{\widehat{LIS}_{i}(\bd{x})\}_{i=1}^m$.
Due to the identity
\[
FDR=E\left[\frac{1}{R\vee 1}\sum_{i=1}^R LIS_{(i)}(\bd{x})\right],
\]
the LIS-based testing procedure in~\eqref{def: LIS procedure} is valid for controlling FDR at level $\alpha$.
Under mild conditions,
this procedure is asymptotically 
optimal in minimizing the FNR~\citep{Sun09,xie2011optimal,Shu15}. 
The LIS theory of \citet{Sun09} is applicable to spatial models that satisfy a monotone ratio condition (MRC)
(their equation (3)).
While their article primarily illustrates the theory through hidden Markov models (HMM), it also acknowledges the broad applicability of the MRC. The theory is extendable to a generalized MRC in \citet{Shu15} (their equation (B.1)). Thus, one needs the generalized MRC rather than HMM to apply the LIS theory.

\begin{figure*}[t!] 
\centering
\includegraphics[width=0.7\textwidth]{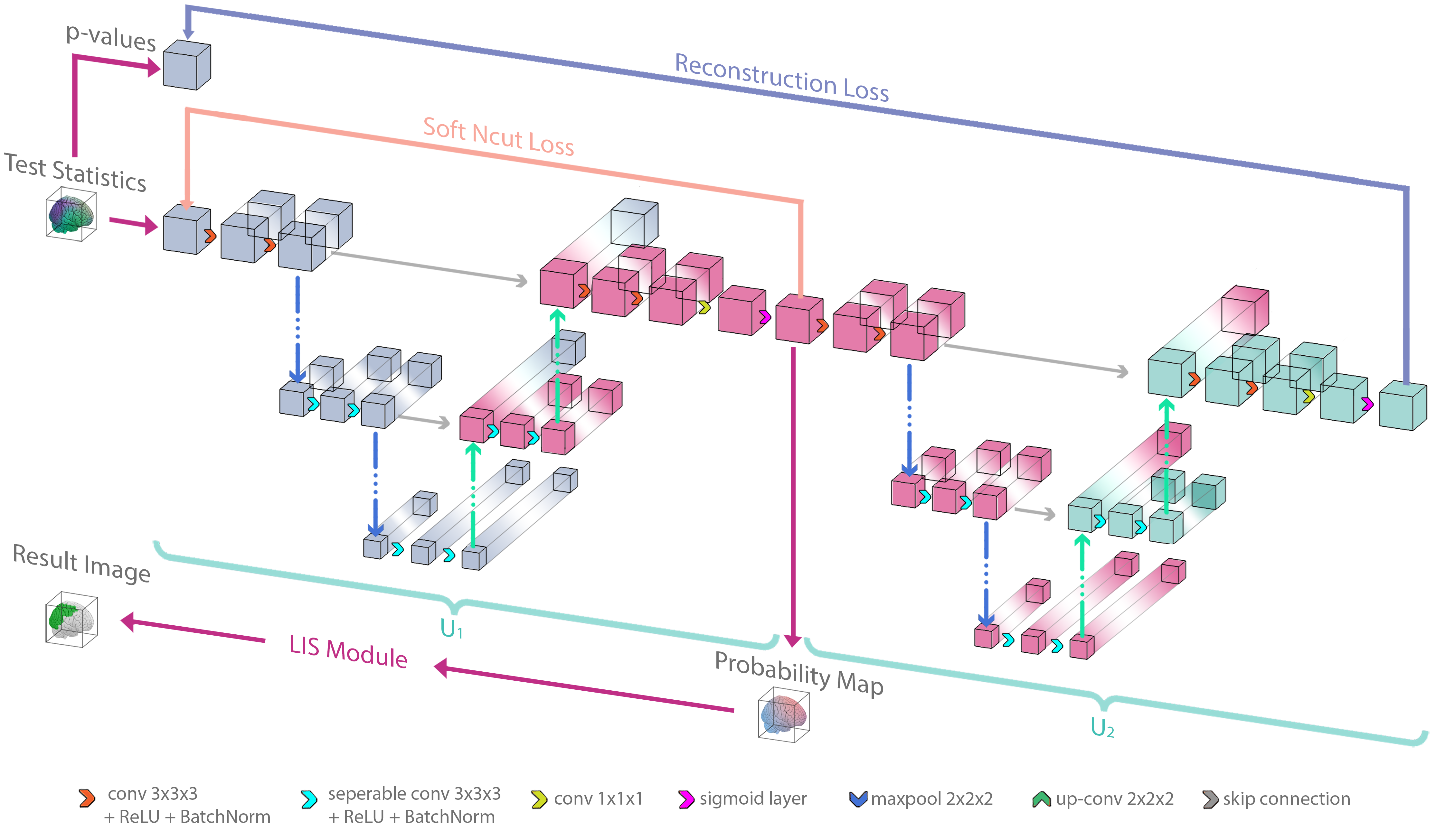}
\vspace{-0.5cm}
\caption{The network architecture of DeepFDR.}
\label{fig:model}\vspace{-0.3cm}
\end{figure*}

\vspace{-0.3cm}
\subsection{DeepFDR}
\vspace{-0.3cm}
Most deep learning-based methods
for image segmentation produce
segmentation probability maps 
$\{\{\widehat{P}(s_i=k|\bd{x})\}_{i=1}^m\}_{k=0}^{K-1}$
as the basis for label assignment,
where $\bd{x}$ is the input image for segmentation, 
$s_i$ is the segmentation label at the $i$-th voxel/pixel, and $K$ is the number of label classes. 
We  establish a connection between the image segmentation 
with $K=2$ classes and
voxel-based multiple testing by letting the input image for segmentation $\bd{x}$ be the 3D map of test statistics and assuming
that segmentation label $s_i=k$ corresponds to 
the null hypothesis state $h_i=k$ for $k=0,1$.
Consequently, the segmentation probability map $\{\widehat{P}(s_i=0|\bd{x})\}_{i=1}^m$ may serve as
an estimate of
the LIS map
$\{LIS_i(\bd{x})=P(h_i=0|\bd{x})\}_{i=1}^m$.
This insight motivates us to adopt a deep learning-based image segmentation method for voxel-based multiple testing.
As mentioned in Section~\ref{sec: intro},
only unsupervised image segmentation methods 
are potentially suitable for our multiple testing problem.
Particularly,
the W-net~\citep{xia2017wnet}
is unsupervised and also 
generates the segmentation probability map.
Moreover, the U-net structure used by the W-net
excels at  capturing multi-scale 
spatial information,
effectively addressing short and long-range spatial dependencies as well as spatial heterogeneity.
Thus,
we choose to adopt the W-net
and make slight modifications for multiple testing purposes.
We then use its
segmentation probability map
 as
an estimate of
the LIS map
for the LIS-based testing procedure given in~\eqref{def: LIS procedure}.

Figure~\ref{fig:model}
provides an overview of our DeepFDR architecture, which is based on the W-net. The input data for the network include the 3D map of test statistics
$\bd{x}=[x_1,\dots,x_m]$ and its corresponding 3D map of p-values $\bd{p}=[p_1,\dots, p_m]$.
The network consists of two cascaded U-nets.
The first U-Net, $\bd{U}_1$, generates the  segmentation probability map
$\{\widehat{P}(s_i=0|\bd{x})\}_{i=1}^m$
using the 
soft normalized cut (Ncut) loss
given in~\eqref{def: soft-Ncut loss}. The second U-Net, $\bd{U}_2$,
reconstructs the p-values $\bd{p}$ from the  segmentation probability map using the mean squared error in~\eqref{def: recon loss} as the reconstruction loss. 
The soft Ncut loss plays a crucial role in partitioning the test statistics $\bd{x}$ into meaningful clusters, akin to the segmentation of an image. The reconstruction loss refines the segmentation probability map by 
enforcing the map to retain sufficient information from the input image. 
The two loss functions are alternately minimized, following the algorithm outlined in 
Algorithm~\ref{alg: DeepFDR}.
This iterative process results in the final segmentation probability map
$\{\widehat{P}(s_i=0|\bd{x})\}_{i=1}^m$. 
Subsequently, this map is fed into 
our LIS module to 
obtain the estimated LIS map 
$\{\widehat{LIS}_i(\bd{x})\}_{i=1}^m$ as per \eqref{eqn: label flip}.
Finally, this LIS map is plugged into the LIS-based testing procedure~\eqref{def: LIS procedure} to yield the multiple testing results.
DeepFDR combines the strengths of deep learning-based image segmentation with the LIS-based testing procedure to effectively handle voxel-based multiple testing tasks.
The key components of the network are elaborated below.

{\bf Soft Ncut loss.} 
We use the soft Ncut loss as the loss function 
for the first U-net $\bd{U}_1$.
The original Ncut loss \citep{shi2000normalized} is widely used in data clustering and image segmentation.
The loss for 
 two classes~is 
\begin{align*}
&Ncut_2(V)=\sum_{k=0}^1\frac{cut(A_k,V\setminus A_k)}{assoc(A_k,V)}\nonumber\\
&=2-\sum_{k=0}^1\frac{assoc(A_k,A_k)}{assoc(A_k,V)}
=2-\sum_{k=0}^1 \frac{\sum_{i\in A_k, j\in A_k}w_{ij}}{\sum_{i\in A_k, j\in V}w_{ij}},
\end{align*}
where $V$ is the set of all voxels, $A_k$ is the set of voxels in class $k$, 
$cut(A,V\setminus A)=\sum_{i\in A, j\in V\setminus A} w_{ij}$ is the total weight of the edges that can be removed between sets $A$ and $V\setminus A$, 
and $assoc(A,B)=\sum_{i\in A, j\in B} w_{ij}$ is the total weight of edges connecting voxels in set $A$ to all voxels in set $B$. Minimizing the Ncut loss can simultaneously minimize the total normalized disassociation between classes and maximize the total normalized association within classes.  
To obtain the sets $A_0$ and $A_1$, 
the argmax function 
is used to assign
the label $k_i^*=\arg\max_{k\in\{0,1\}}\widehat{P}(s_i=k|\bd{x})$ 
 to each $i$-th voxel.
To avoid the nondifferetiable argmax function in computing the Ncut loss, \cite{xia2017wnet} proposed the soft Ncut loss, 
which is differentiable,
by using 
the soft labels 
$\{\{\widehat{P}(s_i=k|\bd{x})\}_{i=1}^m\}_{k=0}^{1}$
instead of the hard labels $\{k_i^*\}_{i=1}^m$.
This allows the loss to be minimized using gradient descent algorithms for the W-net. The soft Ncut loss for two classes is defined~as
\begin{align}\label{def: soft-Ncut loss}
&L_{\text{soft-Ncut}}(\bd{\theta}_1)\\
&= 2- \sum_{k=0}^1\frac{ \sum_{1\le i,j\le m }w_{ij}\widehat{P}(s_i=k|\bd{x})\widehat{P}(s_j=k|\bd{x})}{\sum_{1\le i,j\le m} w_{ij}\widehat{P}(s_i=k|\bd{x})},\nonumber 
\end{align}
where 
\[\label{eq: P_hat=U_1}
[\widehat{P}(s_i=0|\bd{x})]_{i=1}^m
=[1-\widehat{P}(s_i=1|\bd{x})]_{i=1}^m
=\bd{U}_1(\bd{x};\bd{\theta}_1)
\]
is the segmentation probability map
obtained from the first U-net $\bd{U}_1$ with  parameters $\bd{\theta}_1$,
the weight 
\[
w_{ij}=\exp\left(-\frac{|x_i-x_j|^2}{\sigma_x^2}-\frac{\|\bd{\ell}_i-\bd{\ell}_j\|_2^2}{\sigma_\ell^2}\right)I(\|\bd{\ell}_i-\bd{\ell}_j\|_\infty\le r),
\]
with 
$(\sigma_x,\sigma_\ell,r)=(11,3,3)$ in our paper,
$\bd{\ell}_i$ contains the 3D coordinates, and $I(\cdot)$ is the indicator function.

{\bf Reconstruction loss.}
We use the mean squared error as the 
reconstruction loss for the
second U-net $\bd{U}_2$:
\be\label{def: recon loss}
L_{\text{recon}}(\bd{\theta}_1,\bd{\theta}_2) = \frac{1}{m}\sum_{i=1}^{m}(p_i-\widehat{p}_i)^2
\ee
where 
\vspace{-0.77cm}
\begin{align*}
\widehat{\bd{p}}=[\widehat{p}_i]_{i=1}^m
&=\bd{U}_2([\widehat{P}(s_j=0|\bd{x})]_{j=1}^m;\bd{\theta}_2)\\
&=\bd{U}_2(\bd{U}_1(\bd{x};\bd{\theta}_1);\bd{\theta}_2)
\end{align*}
are the reconstructed p-values 
obtained from the second U-net $\bd{U}_2$ with parameters $\bd{\theta}_2$.
Unlike the original W-net,  we use the p-values $\bd{p}$ for reconstruction rather than the target image $\bd{x}$, which is the map of test statistics in our context.
This modification is made because the reconstructed p-values $\widehat{\bd{p}}$ can be effectively 
constrained within the range [0,1] using a sigmoid layer. In contrast, 
if we were to use the reconstructed test statistics $\widehat{\bd{x}}$, they might not
have a well-defined range if the original $\bd{x}$ (e.g., t-statistics) lacks one. Our initial simulation study also indicated
that using p-values for reconstruction yields superior results.
Parameters $\bd{\theta}_1$ and $\bd{\theta}_2$ are simultaneously updated in the minimization of the reconstruction loss.

{\bf LIS module and label flipping.} 
The LIS module is a novel addition to the W-net architecture, enabling the implementation of the LIS-based testing procedure~\eqref{def: LIS procedure}. 
Note that the final segmentation probability map
$\{\widehat{P}(s_i=0|\bd{x})\}_{i=1}^m$ from $\bd{U}_1$ cannot be directly used as
the estimated LIS map $\{\widehat{LIS}_i(\bd{x})\}_{i=1}^m$.
Since the segmentation process here is unsupervised without ground-truth labels, the segmentation label classes may be arbitrarily encoded as 
``0" and ``1", potentially not corresponding well to the hypothesis state label classes. For example, it is possible that segmentation label $s_i=1$ ($s_i=0$) corresponds to hypothesis state label
$h_i=0$ ($h_i=1$, resp.). 
To address this issue, we perform label flipping to correct the possible discrepancy. 
We compare the sets of significant voxels discovered by the LIS-based testing procedure based on  $\widehat{LIS}_{i}(\bd{x})=\widehat{P}(s_i=0|\bd{x})$ and $\widehat{LIS}_{i}(\bd{x})=\widehat{P}(s_i=1|\bd{x})$, respectively, denoted as $S_{\widehat{P}0}(\alpha)$ and $S_{\widehat{P}_1}(\alpha)$, with the discovery set $S_Q(\alpha)$ obtained using the q-value method.
Since  approximately $100(1 - \alpha)\%$ of voxels in $S_Q(\alpha)$
are true signals
due to the robust FDR control of
the q-value method,
our DeepFDR’s discovery set is expected to encompass the majority of voxels in $S_Q(\alpha)$.
Here, we use $S_Q(\alpha)$ as the reference set,
owing to 
the q-value method's superior performance
over BH and LocalFDR methods
and its faster computation
than other spatial FDR methods as shown in our simulation.
We apply the widely-used Dice similarity coefficient~\citep{dice1945measures} 
to measure the similarity 
between $S_{\widehat{P}_0}(\alpha)$ or $S_{\widehat{P}_1}(\alpha)$
and $S_Q(\alpha)$. 
The Dice coefficient
for any two sets $A$ and $B$
is defined as the normalized size of 
their intersection:
\[
Dice(A,B)=\frac{2|A\cap B|}{|A|+|B|}.
\]
If $Dice(S_{\widehat{P}_0}(\alpha),S_Q(\alpha))<Dice(S_{\widehat{P}_1}(\alpha),S_Q(\alpha))$, we flip the segmentation label classes.
Equivalently,
the label flipping is performed as follows:
\begin{align}\label{eqn: label flip}
&\widehat{LIS}_i(\bd{x})\overset{\text{def}}{=}\widehat{P}(h_i=0|\bd{x})\\
&=
\begin{cases}
\widehat{P}(s_i=0|\bd{x}), &\text{if}~Dice(S_{\widehat{P}_0}(\alpha),S_Q(\alpha))\\
&\qquad\ge Dice(S_{\widehat{P}_1}(\alpha),S_Q(\alpha));\\
\widehat{P}(s_i=1|\bd{x}), &\text{otherwise}.\nonumber
\end{cases}
\end{align}
If the q-value method yields no
or a very small number of discoveries,
one may gradually increase the nominal 
FDR level $\alpha_Q\ge \alpha$ exclusively for the q-value method to obtain an acceptable  $S_Q(\alpha_Q)$, and then apply the criterion~\eqref{eqn: label flip}.
If $|S_Q(\alpha_Q)|$ remains very small despite a significant increase in $\alpha_Q$ compared to the original $\alpha$, 
one may consider using p-values instead. For example, 
gradually decrease the uncorrected significance level $\alpha_P\le \alpha$
for p-values, and in \eqref{eqn: label flip} replace
$S_Q(\alpha_Q)$ with 
$S_P(\alpha_P)$, which is the set of voxels with p-values  $<\alpha_P$.
It is important to assume that the uncorrected p-value rejection set $S_P(\alpha)$ at level $\alpha$ is not excessively small; otherwise, one may need to contemplate increasing the nominal  FDR level $\alpha$ for the multiple testing problem.

\vspace{-0.1cm}

\begin{algorithm}
\caption{Algorithm for DeepFDR}\label{alg: DeepFDR}
\begin{algorithmic}[1]
\INPUT 3D volumes of test statistics $\bd{x}$ and p-values~$\bd{p}$, and
prespecified FDR level $\alpha$.
\For{epoch $t=1:T$}
    \State Only update parameter $\bd{\theta}_1$ by minimizing the $L_{\text{soft-Ncut}}$ in~\eqref{def: soft-Ncut loss};
    \State Update both parameters $\bd{\theta}_1$ and $\bd{\theta}_2$
    by minimizing the $L_{\text{recon}}$ in~\eqref{def: recon loss};
\EndFor
 \State Compute the LIS estimates $\{\widehat{LIS}_i(\bd{x})\}_{i=1}^m$ by~\eqref{eqn: label flip};
\State  Conduct the LIS-based testing procedure~\eqref{def: LIS procedure} with $\{\widehat{LIS}_i(\bd{x})\}_{i=1}^m$;
\OUTPUT A 3D volume of estimates for the null hypothesis states $\bd{h}$.
\end{algorithmic}
\end{algorithm}

\vspace{-0.1cm}

{\bf Detailed network architecture.}
Our DeepFDR network architecture, as depicted in Figure~\ref{fig:model}, is primarily based on the structure of the W-net~\citep{xia2017wnet}. It comprises two cascaded U-nets, each featuring a contracting path and an expanding path that span three levels of network layers. 
The network is equipped with a total of  10 pairs of two consecutive $3\times 3\times 3$ convolution layers,
which have 64, 128, and 256 feature channels at the top, middle, and bottom levels, respectively.
Each of these convolution layers is followed by a rectified linear unit (ReLU; \cite{nair2010rectified}) and batch normalization~\citep{ioffe2015batch}. 
While regular convolutions are utilized at the top level,
depthwise separable convolutions~\citep{chollet2017xception} are employed at the other two levels to significantly reduce parameters.
The feature maps are downsampled from upper levels to lower levels by a $2\times 2\times 2$ max-pooling operation with a stride of 2 to halve spatial dimensions,
but they are upsampled from lower levels to upper levels by a $2\times 2\times 2$ transposed convolution with a stride of 2 to double spatial dimensions.
Skip connections are used to concatenate the feature maps
in the contracting path with those in the expanding path to capture the multi-scale spatial information.
Within each U-net, the last two layers consist of a
$1\times 1\times 1$ convolution layer and a sigmoid layer.
The convolution layer transforms all feature maps into a single feature map,
enabling the subsequent sigmoid layer
to generate the segmentation probability map $\{\widehat{P}(s_i=0|\bd{x})\}_{i=1}^m$ for $\bd{U}_1$ 
or the reconstructed p-value map $\widehat{\bd{p}}$ for $\bd{U}_2$.
The segmentation probability map $\{\widehat{P}(s_i=0|\bd{x})\}_{i=1}^m$ from $\bd{U}_1$
and the input test statistics $\bd{x}$ are used to minimize the soft Ncut loss given in~\eqref{def: soft-Ncut loss} with parameter $\bd{\theta}_1$,
and the reconstructed and original p-value maps $\widehat{\bd{p}}$ and $\bd{p}$ are used to minimize
the reconstruction loss given in~\eqref{def: recon loss} with parameters $\bd{\theta}_1$ and $\bd{\theta}_2$.

{\bf Network training.}
In contrast to supervised deep learning models
which have access to multiple images with predefined ground-truth labels for training and validation,
voxel-based multiple testing, as an unsupervised-learning problem,
 only has a single image of the test statistics $\bd{x}$ 
and thus has no straightforward validation set,
and moreover lacks very effective validation criteria 
due to the absence of predefined ground-truth labels.
While one might consider splitting the image of $\bd{x}$ into patches, this approach would lose long-range spatial structures
and ignore the spatial heterogeneity. 
Alternatively, one could divide the sample data (e.g., subjects' FDG-PET images) into two parts and compute their 
respective maps of test statistics for training and validation, 
but the reduced sample size leads to less powerful test statistics.
In our method,
we utilize the complete map of test statistics from all sample data
as the training image,
and do not allocate an image for validation according to the W-net paper~\citep{xia2017wnet}.
Instead, multiple regularization techniques are applied to prevent overfitting~\citep{buhmann1999unsupervised} and enhance training stability:
a dropout~\citep{srivastava2014dropout} of rate 0.5 before the second max-pooling of each U-net, 
weight decay~\citep{krogh1991simple} of rate $10^{-5}$ in the stochastic gradient descent (SGD) optimizer, 
batch normalization~\citep{ioffe2015batch} after each ReLU layer,
and early stopping~\citep{prechelt2002early} based on the two loss functions.
Algorithm~\ref{alg: DeepFDR} outlines our DeepFDR algorithm, which alternately optimizes the two loss functions. At each epoch,
 the algorithm updates the parameter $\bd{\theta}_1$ for $\bd{U}_1$  
by minimizing the $L_{\text{soft-Ncut}}$ loss  in~\eqref{def: soft-Ncut loss},
and 
then simultaneously updates the parameters $\bd{\theta}_1$ and $\bd{\theta}_2$ for $\bd{U}_1$ and $\bd{U}_2$
by minimizing the
$L_{\text{recon}}$ loss  in~\eqref{def: recon loss}.
After network training, 
the final segmentation probability map
is generated using the trained network with dropout disabled,
and 
is then passed through our LIS module to 
obtain the estimated LIS map 
$\{\widehat{LIS}_i(\bd{x})\}_{i=1}^m$ by~\eqref{eqn: label flip}.
This estimated LIS map is plugged into
 the LIS-based testing procedure~\eqref{def: LIS procedure} to yield the multiple testing result.

\vspace{-0.3cm}
\section{NUMERICAL RESULTS}
\vspace{-0.3cm}
We compare our DeepFDR with classic and recent FDR control methods in Section~\ref{sec: simulation}
through simulations  and 
in Section~\ref{sec: real data}
using FDG-PET data
from the Alzheimer’s Disease Neuroimaging Initiative (ADNI).

\vspace{-0.3cm}
\subsection{Methods for Comparison}
\vspace{-0.3cm}
We conducted a comparative evaluation of  our DeepFDR against
eight existing methods, including
BH~\citep{Benj95}, q-value~\citep{Stor03}, LocalFDR~\citep{efron2004large}, HMRF-LIS \citep{Shu15}, SmoothFDR~\citep{Tans18}, LAWS~\citep{laws}, NeuralFDR~\citep{xia2017neuralfdr}, and OrderShapeEM (OSEM; \citet{Cao_Chen_Zhang_2022}).
The BH, q-value, and LocalFDR methods are 
classic FDR control methods 
developed for independent tests, but
HMRF-LIS, SmoothFDR, and LAWS 
are state-of-the-art spatial methods
applicable to 3D image data.
HMRF-LIS uses 1-nearest-neighbor HMRFs to model spatial dependencies and then applies the LIS-based testing procedure. 
SmoothFDR utilizes an empirical-Bayes approach to enforce spatial smoothness with lasso to detect localized regions of significant test statistics.
LAWS constructs structure-adaptive weights
based on the estimated local sparsity levels
to weigh p-values.
NeuralFDR is designed for
multiple testing problems with covariates available,
and employs a deep neural network to learn the
p-value threshold as a function of the covariates; in our context,
we used the 3D coordinates as three covariates for NeuralFDR. OSEM extends the LocalFDR method  by incorporating auxiliary information on the order of prior null probabilities, which is often lacking in voxel-based multiple testing; to serve as the auxiliary information, q-values were employed in simulations, and both q-values and BH-adjusted p-values were attempted in the real-data analysis. The detailed implementations of the nine methods are given in Appendix. 

\vspace{-0.3cm}
\subsection{Simulation Studies}\label{sec: simulation}
\vspace{-0.3cm}
{\bf Simulation settings.}
We generated each simulated dataset on a  lattice cube
with size $m=30\times 30\times 30$. 
The ground-truth hypothesis state labels $\bd{h}=[h_1,\dots, h_m]$ were generated based on the ADNI FDG-PET dataset in Section~\ref{sec: real data}. Specifically, 
we used the result of 
the q-value method
with nominal FDR level 0.01 for the comparison between the early mild cognitive impairment group and the cognitively normal group;
three $30\times 30\times 30$ lattice cubes were randomly cropped from the brain volume of the q-value result, respectively with about 10\%, 20\%, and 30\% 
of voxels tested as significant;
in the three cubes,
we set ground-truth values of $h_i=1$ for the significant voxels 
and $h_i=0$ for the remaining voxels.
For each cube,
the test statistics $\bd{x}=[x_1,\dots, x_m]$ 
were generated 
using the Gaussian mixture
model:
$x_i|h_i\sim (1-h_i)N(0,1)+h_i\{\frac{1}{2}N(\mu_1,\sigma_1^2)+\frac{1}{2}N(2,1)\}$.
We varied $\mu_1$ from $-4$ to $0$ with fixed $\sigma_1^2=1$, and varied $\sigma_1^2$ from 0.125 to 8
with fixed $\mu_1=-2$.
In total, we generated 45 simulation settings, including 15 different combinations of $(\mu_1,\sigma_1^2)$ for each of the three cubes with different proportions of signals.
We conducted the nine FDR control methods
with a nominal FDR level $\alpha=0.1$
for 50 independent replications of each simulation setting.
FDR, FNR, the average number of true positives (ATP), and computational time
for each method 
were computed based on the 50 replications.

\begin{figure*}[h!] 
\centering
\includegraphics[width=0.85\textwidth]{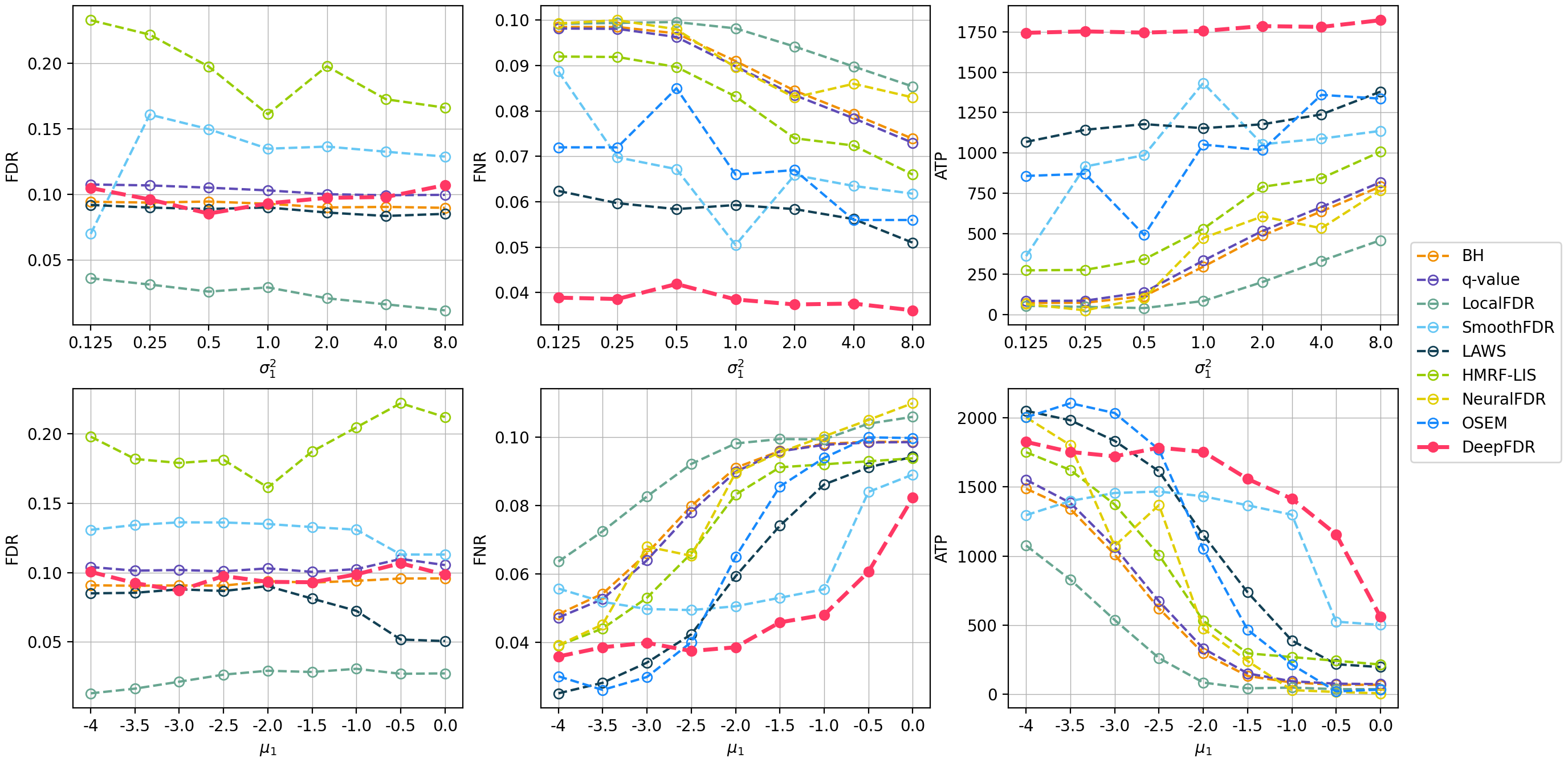}
\vspace{-0.4cm}
\caption{Simulation results for the cube with $P_1\approx 10\%$. All FDRs of NeuralFDR and almost all FDRs of OSEM are too large, and thus their FDRs are not shown in this figure; see Figure~\ref{fig:simulation_0.1}, instead.}
\label{fig: siml p=0.1}\vspace{-0.1cm}
\end{figure*}

\vspace{-0.05cm}
{\bf Multiple-testing results.}
Figures~\ref{fig: siml p=0.1} and \ref{fig: siml p=0.2}-\ref{fig:simulation_0.3} display the multiple-testing results 
for the three cubes with signal proportion (denoted by $P_1$)
approximately equal to 10\%, 20\%, and 30\%, respectively. 
We see that our DeepFDR 
well controls the FDR around the nominal level 0.1, and performs the best in 39 simulation settings and ranks second in the other 6 settings in terms of smallest FNR, largest ATP and controlled FDR.
In particular, 
for weak signal cases where
$\mu_1\in[-2,0]$ and $\sigma_1^2=1$,
DeepFDR surpasses the other valid FDR control methods by a large margin. 
For strong signal cases with $\mu_1\in\{-4,-3.5,-3\}$ and $\sigma_1^2=1$ when
$P_1\approx 10\%$ or $30\%$,
DeepFDR is outperformed by LAWS;
this behavior is reasonable since the optimality of DeepFDR's LIS-based testing procedure is asymptotic and subject to certain conditions~\citep{Sun09,xie2011optimal}.
It is observed that
all FDRs of NeuralFDR are 
more than 0.2 larger than the nominal level 0.1,
OSEM and HMRF-LIS are not valid in FDR control for almost all simulation settings,
and SmoothFDR is not valid
for almost all settings with $P_1\approx 10\%$ and 
some settings with $P_1\approx 20\%$.
This may be owing to 
the incompatible assumption made by NeuralFDR for spatial data (see Section~\ref{sec: intro}),
the failure of OSEM to consider spatial dependence, 
the inadequate spatial modeling by  HMRF-LIS, 
and the oversmoothing effect of  SmoothFDR.
The figures show that
BH, LocalFDR, and LAWS are often
conservative in FDR control with FDR smaller than the nominal level with a large distance.
The q-value method well controls FDR around 0.1, and has smaller FNR and larger ATP than
BH and LocalFDR, but is inferior to the spatial methods LAWS and DeepFDR.

\vspace{-0.05cm}
{\bf Timing performance.}
DeepFDR, NeuralFDR, and HMRF-LIS were executed on a NVIDIA RTX8000 GPU (48GB memory), and the other six methods were run on a server with 20 Intel Xeon Platinum 8268 CPUs (2.90GHz, 64GB memory).
The computational time was computed based on the simulation setting with $(\mu_1,\sigma_1^2)=(-2,1)$ and $P_1\approx 20\%$.
Table~\ref{table:run time for simulation} presents the mean and standard deviation (SD) of the runtime over the 50 simulation  replications.
Given that BH, q-value, and LocalFDR methods are designed for independent tests rather than spatial data, it is not surprising that they exhibit the fastest performance, each completing with a mean runtime of less than 5 seconds. 
Our DeepFDR 
 boasts a mean runtime of 7.21 seconds, with an SD of 1.22 seconds,
 which is approximately 1.7 times the runtime of the q-value method. 
However, it remains notably faster than the other four methods,
 requiring only about 
 1/2 of the time used by OSEM, 
 1/8 of HMRF-LIS, 1/20 of SmoothFDR, 1/50 of LAWS, and  1/860 of NeuralFDR.

\vspace{-0.3cm}
\subsection{Real-data Analysis}\label{sec: real data}
\vspace{-0.2cm}
FDG-PET is a widely used imaging technique in early diagnosis and monitoring progression of Alzheimer's disease (AD). This technique assesses brain glucose metabolism, which typically decreases in AD cases. The difference in brain glucose metabolism between two population groups can be investigated by 
testing the difference of their voxel-level
population means in the SUVR from FDG-PET, leading to a high-dimensional spatial multiple testing problem. 
We employed voxel-based multiple testing methods
to compare the mean SUVR difference between
the cognitively normal (CN) group
and each of the following three groups:
early mild cognitive impairment patients
with conversion to AD (EMCI2AD),
late mild cognitive impairment patients
with conversion to AD (LMCI2AD),
and the AD group.

{\bf ADNI FDG-PET dataset.} The FDG-PET image dataset used in this study was obtained from the ADNI database (adni.loni.usc.edu).
The ADNI was launched in
2003 as a public-private partnership, led by Principal Investigator Michael W. Weiner,
MD. The primary goal of ADNI has been to test whether serial magnetic resonance imaging, positron emission tomography, other biological markers, and clinical and
neuropsychological assessment can be combined to measure the progression of mild
cognitive impairment and early AD. 
The dataset consists of  baseline FDG-PET images 
from 742 subjects, including 286 CN subjects, 42 EMCI2AD patients, 175 LMCI2AD patients, and 239 AD patients. All 742 FDG-PET images were preprocessed using the Clinica software~\citep{routier2021clinica}
to ensure spatial normalization to the MNI IXI549Space template and intensity normalization based on the average uptake value in the pons region. 
We considered the 120 brain regions of interest (ROIs) from the AAL2 altas~\citep{rolls2015implementation}.
The total number of voxels in the 120 ROIs is 439,758,
and the number of voxels in each ROI ranges from 107 to 12,201 with a median of 2874 (see Table~\ref{table:roi voxel number}).
For each ROI voxel, we ran
a linear regression 
with the voxel's SUVR as the response variable
and 
the dummy variables of the EMCI2AD, LMCI2AD, and AD 
groups as explanatory variables (where CN was used as the reference group),
adjusting for patient's age, gender, race, ethnicity, education, marital status, and APOE4 status. 
The voxel-level t-statistics 
for regression coefficients of the three groups' dummy variables and associated p-values
were thus obtained
for the three comparisons: EMCI2AD vs. CN, LMCI2AD vs. CN, and AD vs. CN.
Z-statistics were transformed from t-statistics for certain FDR control methods that require them as input.

{\bf Multiple-testing results.} 
All FDR control methods were conducted with the nominal FDR level $\alpha=0.001$ for each of the three comparisons on the 439,758 ROI voxels.
OSEM finds no discoveries in the three comparisons when using q-values
or BH-adjusted p-values as its  auxiliary
information.
Figures~\ref{fig: zvalue EMCI2AD}--\ref{fig: zvalue AD}
present the discoveries obtained by each  method.
For all methods except SmoothFDR and OSEM,
it is observed that
most discovered brain areas 
exhibit hypometabolism, 
and the affected areas expand 
and deteriorate
during the AD progression from CN to EMCI2AD, then to LMCI2AD, and finally to AD.
Figures~\ref{fig:HP EMCI}--\ref{fig:HP AD} 
 show the proportion of discoveries 
found  by each method in each ROI for the three comparisons. 
The proportion of discoveries generally increases in each ROI during the AD progression, 
again indicating the growing impact of the disease on the brain. 

In the AD vs. CN comparison, as shown in Figures~\ref{fig: zvalue AD} and \ref{fig:HP AD}, 
all methods, except OSEM, SmoothFDR and NeuralFDR, exhibit similar distributions for the  proportion of discoveries over the 120 ROIs.
SmoothFDR and NeuralFDR appear to overestimate signals, as a significant amount of their discoveries have p-values exceeding 0.001, 0.01, and 0.05 thresholds.
Specifically, for SmoothFDR, NeuralFDR, and our DeepFDR, among their respective discoveries,
35.1\%, 47.2\%, and 2.6\% have p-values $>0.001$,
22.9\%, 37.2\%, and 0.094\% have p-values $>0.01$, and
12.1\%, 28.5\%, and 0.0096\% have p-values $>0.05$.
For the LMCI2AD vs. CN comparison, as shown in Figures~\ref{fig: zvalue LMCI2AD} and \ref{fig:HP LMCI}, 
the non-spatial methods BH, q-value, and LocalFDR are conservative in discoveries,
spatial methods HMRF-LIS, LAWS, and our DeepFDR exhibit similar distributions of their discoveries,
while SmoothFDR and NeuralFDR continue to demonstrate an overestimation of signals. 
Among the respective discoveries of SmoothFDR, NeuralFDR, and our DeepFDR, 
53.6\%, 66.1\%, and 5.3\% have p-values $>0.001$,
31.2\%, 52.5\%, and 0.027\% have p-values $>0.01$, and
18.4\%, 42.1\%, and 0\% have p-values $>0.05$.
This highlights the challenge of effectively controlling FDR for SmoothFDR and NeuralFDR, 
whereas our DeepFDR presents credible discoveries with significantly smaller p-values in the two comparisons.
Note that the nominal FDR level $\alpha$ is $0.001$, but
it does not necessarily imply that a discovery with p-value slightly above 0.001 is definitively not a signal,
because such thresholding of p-values  does not account for the spatial dependence in neuroimaging data.
However, if a discovery has a p-value  much larger than the nominal level 0.001, e.g., 0.05, 
it is more likely to be a false discovery.

The EMCI2AD vs. CN comparison is particularly challenging among the three comparisons, yet it holds significant promise for  early detection of AD. 
In this comparison, 
BH, q-value, LocalFDR, LAWS, and OSEM fail to 
yield any discoveries,
and HMRF-LIS identifies only 3 discoveries, 
Indeed, there are only 101 voxels with p-values $< 0.001$, which reflects the difficulty of this comparison.
NeuralFDR finds 14,342 discoveries, which
are scattered across the brain as shown in Figures~\ref{fig: zvalue EMCI2AD} and \ref{fig:HP EMCI}. 
SmoothFDR identifies 86,719 discoveries,
but the result appears oversmoothed as shown in Figure~\ref{fig: zvalue EMCI2AD}.
NeuralFDR and SmoothFDR seem to overestimate the signals,
with 95.7\% and 68.1\% of their discoveries
having p-values $>0.05$. In contrast,
DeepFDR provides 1087 discoveries, of which 82 are among the 101 voxels with p-values $< 0.001$.
Impressively,
88.9\%, 99.3\%, and 100\% of DeepFDR's discoveries have p-values less than 0.005, 0.01, and 0.05, respectively.
All of DeepFDR's discoveries  are
located in the left hemisphere, with
1080 of them found
in left parahippocampal gyrus ($n{=}276, P{=}11.85\%$), left hippocampus ($n{=}130, P{=}5.84\%$), left inferior temporal gyrus ($n{=}392, P{=}5.18\%$),  left middle temporal gyrus ($n{=}244, P{=}2.08\%$), and left fusiform gyrus ($n{=}38, P{=}0.70\%$). 
This aligns with prior research  suggesting greater vulnerability of the left hemisphere to AD~\citep{thompson2001cortical,thompson2003dynamics,roe2021asymmetric}. 
These five ROIs are known to be early affected by AD~\citep{echavarri2011atrophy, braak1993staging, convit2000atrophy}, providing additional support for the validity of DeepFDR's discoveries.

{\bf Timing performance.} 
We executed the methods using the same computational resource as specified  in Section~\ref{sec: simulation}.
Table~\ref{table:run time for simulation}  shows the mean and SD of the runtime over the three comparisons for the ADNI FDG-PET data.
The three non-spatial methods BH, q-value and LocalFDR exhibit  dominant performance. 
Our DeepFDR follows closely in efficiency; it averaged a runtime of of 89.98 seconds with an SD of
5.17 seconds,
which is merely 1.31 times the runtime of the q-value method.
In stark contrast, 
the mean runtime for each of the other five methods exceeds 5 hours, with LAWS taking nearly 7 days.
These results emphasize the high computational efficiency of our DeepFDR when tackling the voxel-based multiple testing challenge in neuroimaging data analysis.

\vspace{-0.3cm}
\section{CONCLUSION}
\vspace{-0.3cm}
This paper 
proposes DeepFDR, a novel deep
learning-based FDR control method for voxel-based
multiple testing. 
DeepFDR harnesses deep learning-based unsupervised image segmentation, specifically a modified W-net, to effectively capture
spatial dependencies among voxel-based tests, 
and then utilizes the LIS-based testing procedure
to achieve FDR control and minimize the FNR. 
Our extensive numerical studies, including comprehensive simulations and in-depth analysis of 3D FDG-PET images related to Alzheimer's disease, corroborate DeepFDR's superiority over existing methods. DeepFDR consistently demonstrates its ability to effectively control the FDR while substantially reducing the FNR, thereby enhancing the overall reliability of results in neuroimaging studies.
Furthermore, DeepFDR distinguishes itself by its remarkable computational efficiency. By leveraging well-established software and advanced optimization algorithms from the field of deep learning, it stands as an exceptionally fast and efficient solution for addressing  the voxel-based multiple testing problem in large-scale neuroimaging data analysis.

\subsubsection*{Acknowledgements}

Dr. Shu's research was partially supported by the grant R21AG070303 from the National
Institutes of Health (NIH). 
Dr. de Leon's research was partially supported by the NIH grants AG022374, AG12101, AG13616, AG057570, AG057848, and AG058913.
The content is solely the responsibility of the authors and does not necessarily represent the official views of the NIH.

This work was supported in part through the NYU IT High Performance Computing resources, services, and staff expertise.

\begin{enumerate*}[label=*]
\item\label{datause}
Data used in preparation of this article were obtained from the Alzheimer’s Disease
Neuroimaging Initiative (ADNI) database (adni.loni.usc.edu). As such, the investigators
within the ADNI contributed to the design and implementation of ADNI and/or provided data
but did not participate in analysis or writing of this report. A complete listing of ADNI
investigators can be found at:
http://adni.loni.usc.edu/wp-content/uploads/how\_to\_
\\
apply/ADNI\_Acknowledgement\_List.pdf.

\end{enumerate*}

Data collection and sharing for this project was funded by the Alzheimer's Disease
Neuroimaging Initiative (ADNI) (National Institutes of Health Grant U01 AG024904) and
DOD ADNI (Department of Defense award number W81XWH-12-2-0012). ADNI is funded
by the National Institute on Aging, the National Institute of Biomedical Imaging and
Bioengineering, and through generous contributions from the following: AbbVie, Alzheimer’s
Association; Alzheimer’s Drug Discovery Foundation; Araclon Biotech; BioClinica, Inc.;
Biogen; Bristol-Myers Squibb Company; CereSpir, Inc.; Cogstate; Eisai Inc.; Elan
Pharmaceuticals, Inc.; Eli Lilly and Company; EuroImmun; F. Hoffmann-La Roche Ltd and
its affiliated company Genentech, Inc.; Fujirebio; GE Healthcare; IXICO Ltd.; Janssen
Alzheimer Immunotherapy Research \& Development, LLC.; Johnson \& Johnson
Pharmaceutical Research \& Development LLC.; Lumosity; Lundbeck; Merck \& Co., Inc.;
Meso Scale Diagnostics, LLC.; NeuroRx Research; Neurotrack Technologies; Novartis
Pharmaceuticals Corporation; Pfizer Inc.; Piramal Imaging; Servier; Takeda Pharmaceutical
Company; and Transition Therapeutics. The Canadian Institutes of Health Research is
providing funds to support ADNI clinical sites in Canada. Private sector contributions are
facilitated by the Foundation for the National Institutes of Health (www.fnih.org). The grantee
organization is the Northern California Institute for Research and Education, and the study is
coordinated by the Alzheimer’s Therapeutic Research Institute at the University of Southern
California. ADNI data are disseminated by the Laboratory for Neuro Imaging at the
University of Southern California.

\bibliography{AISTATS} 

\section*{Checklist}


 \begin{enumerate}

 \item For all models and algorithms presented, check if you include:
 \begin{enumerate}
   \item A clear description of the mathematical setting, assumptions, algorithm, and/or model. [Yes]
   \item An analysis of the properties and complexity (time, space, sample size) of any algorithm. [Yes]
   \item (Optional) Anonymized source code, with specification of all dependencies, including external libraries. [Yes]
 \end{enumerate}

 \item For any theoretical claim, check if you include:
 \begin{enumerate}
   \item Statements of the full set of assumptions of all theoretical results. [Yes]
   \item Complete proofs of all theoretical results. [Not Applicable]
   \item Clear explanations of any assumptions. [Yes]     
 \end{enumerate}

 \item For all figures and tables that present empirical results, check if you include:
 \begin{enumerate}
   \item The code, data, and instructions needed to reproduce the main experimental results (either in the supplemental material or as a URL). [No. The code is available at https://github.com/kimtae55/DeepFDR. The FDG-PET image dataset used in our paper was obtained from the Alzheimer’s Disease Neuroimaging Initiative (https://adni.loni.usc.edu). We are not allowed to share this dataset. The instructions have been given in the paper and its appendix.]
   \item All the training details (e.g., data splits, hyperparameters, how they were chosen). [Yes]
         \item A clear definition of the specific measure or statistics and error bars (e.g., with respect to the random seed after running experiments multiple times). [Yes]
         \item A description of the computing infrastructure used. (e.g., type of GPUs, internal cluster, or cloud provider). [Yes]
 \end{enumerate}

 \item If you are using existing assets (e.g., code, data, models) or curating/releasing new assets, check if you include:
 \begin{enumerate}
   \item Citations of the creator If your work uses existing assets. [Yes]
   \item The license information of the assets, if applicable. [Yes. If applicable, the license information for
existing codes can be found at the URLs provided in our paper.]
   \item New assets either in the supplemental material or as a URL, if applicable. [Yes]
   \item Information about consent from data providers/curators. [Yes. We have successfully executed the data use agreement with our data provider, the Alzheimer’s Disease Neuroimaging Initiative (ADNI). We have included an acknowledgment to ADNI in this paper.]
   \item Discussion of sensible content if applicable, e.g., personally identifiable information or offensive content. [Not Applicable]
 \end{enumerate}

 \item If you used crowdsourcing or conducted research with human subjects, check if you include:
 \begin{enumerate}
   \item The full text of instructions given to participants and screenshots. [Not Applicable]
   \item Descriptions of potential participant risks, with links to Institutional Review Board (IRB) approvals if applicable. [Not Applicable]
   \item The estimated hourly wage paid to participants and the total amount spent on participant compensation. [Not Applicable]
 \end{enumerate}

 \end{enumerate}

\renewcommand\thesection{A.\arabic{section}}
\setcounter{section}{0}

\renewcommand\theequation{A.\arabic{equation}}
\setcounter{equation}{0}

\renewcommand\thetable{A.\arabic{table}}
\setcounter{table}{0}

\renewcommand\thefigure{A.\arabic{figure}}
\setcounter{figure}{0}

\onecolumn
\section*{APPENDIX}

\subsection*{A.1~~Implementation Details of Comparison Methods}

In this section, we provide a comprehensive overview of the implementation details for all the methods used in our numerical comparisons. 
It is worth noting that the Python versions of the methods consistently demonstrated superior speed compared to their R counterparts. Thus, we prioritized Python versions whenever available, only resorting to R when necessary.
Our numerical studies, including simulations and real-data analysis, were conducted using Python~3.9.7 and R~4.2.1.

{\bf BH and LocalFDR:}
We used the Python package {\bf statsmodels} (v0.12.2) available at 
\url{https://www.statsmodels.org}.

{\bf q-value:} We used 
the Python package {\bf multipy} (v0.16) available at \url{https://github.com/puolival/multipy}.

{\bf HMRF-LIS:} The original implementation is in C++ (\url{https://github.com/shu-hai/FDRhmrf}), but its sequential nature using Gibbs sampling poses scalability challenges. To address this, we have created a Python version that utilizes GPU-based HMRF Gibbs sampling. Although Gibbs sampling is traditionally sequential, the Ising model-based HMRF used by the method exhibits a dependency on neighboring voxels that can be parallelized by modeling the input voxels as a black and white checkerboard. We applied a convolutional operation with a suitable $3\times 3\times 3$ kernel to extract information from neighboring voxels, achieving significant speedup and faster convergence. 
In simulations, we used a single HMRF to model the $30\times 30\times 30$ lattice cube.
But in real-data analysis,
we modeled each ROI with a separate HMRF, following the HMRF-LIS paper.

{\bf SmoothFDR:} We utilized the author-published Python package available at \url{https://github.com/tansey/smoothfdr}, using 20 sweeps. 

{\bf NeuralFDR:} We utilized the author-published Python package available at \url{https://github.com/fxia22/NeuralFDR/tree/master}. The input consists of each test's p-value and corresponding covariates. We used the 3D 
coordinates as the covariates.
We noticed the standard practice of mini-batch based training was not implemented in their code, 
resulting in GPU memory allocation issues when handling the ADNI data. Thus, we modified their code to incorporate mini-batches during the forward pass, aggregating the respective losses instead of inputting the entire training set at once.
For simulations, we used the default parameters in their code,
but in real-data analysis, we set n-init=3 and num-iterations=200 to reduce the computational time. 

{\bf LAWS:} Only R code is available in the Supplementary Materials of its paper at \url{https://doi.org/10.1080/01621459.2020.1859379}. The 3D implementation of LAWS was used in both simulations and real-data analysis. 

{\bf OrderShapeEM (OSEM):} We used the author-published R package available at \url{https://github.com/jchen1981/OrderShapeEM}. 
To serve as the auxiliary information on the order of prior null probabilities,
q-values were employed in simulations, and both q-values and BH-adjusted p-values were attempted in the real-data analysis.

{\bf DeepFDR:} We implemented our algorithm using the Pytorch package (v2.0.1) for the network. The code is available at \url{https://github.com/kimtae55/DeepFDR}.
Most details  can be found in Section 2.3 of our paper.
For training, the SGD optimizer with a momentum of 0.9 and weight decay of
$10^{-5}$ was used with 
Kaiming initialization for weights. 
The learning rate was tuned and early stopping was applied based on the two loss functions.
The best learning rate was 0.05 for most simulation settings and 0.07 for the others,
and is 0.008, 0.001, and 0.006 for EMCI2AD vs CN, LMCI2AD vs CN, and AD vs CN, respectively. 
The algorithm was terminated   before 25 epochs 
for all simulation settings and 10 epochs for all comparisons in real-data analysis.
In a preliminary simulation, 
the parameters $(\sigma_x,\sigma_\ell,r)$ were slightly tuned around the values (10,4,5) used by \cite{xia2017wnet}.
Despite this fine-tuning not significantly altering the results,
these parameters were ultimately set to (11,3,3) for the final simulations and real-data analysis.

BH and q-value methods take a 1D sequence of p-values  as input,
OSEM requires a 1D sequence of p-values and a 1D sequence of auxiliary information on the order of prior null probabilities,
NeuralFDR accepts p-values and  3D coordinates as input, 
LAWS takes a 3D volume of p-values,
LocalFDR requires a 1D sequence of z-values, 
HMFR-LIS and SmoothFDR expect a 3D volume of z-values, and
DeepFDR takes a 3D volume of test statistics (z-values in simulations and t-values in real-data analysis) and the corresponding 3D volume of p-values as input.
In simulations, the 3D volume had a size of $30\times 30\times 30$ given to the other methods,
and DeepFDR zero-padded the volume to size $32\times 32\times 32$ to facilitate the two max-pooling layers in each U-net of its network.   
In real-data analysis, the 3D volume was cropped to size $100\times 120\times 100$ 
from the original brain image size of $121\times 145\times 121$
by removing redundant background voxels;
the non-ROI voxels were set with 0 for t-values and z-values, and 1 for p-values;
only tests on the ROI voxels were used to yield the multiple testing results.

\subsection*{A.2~~Supplementary Tables and Figures for Numerical Results}

\begin{figure*}[h!] 
\centering
\includegraphics[width=0.85\textwidth]{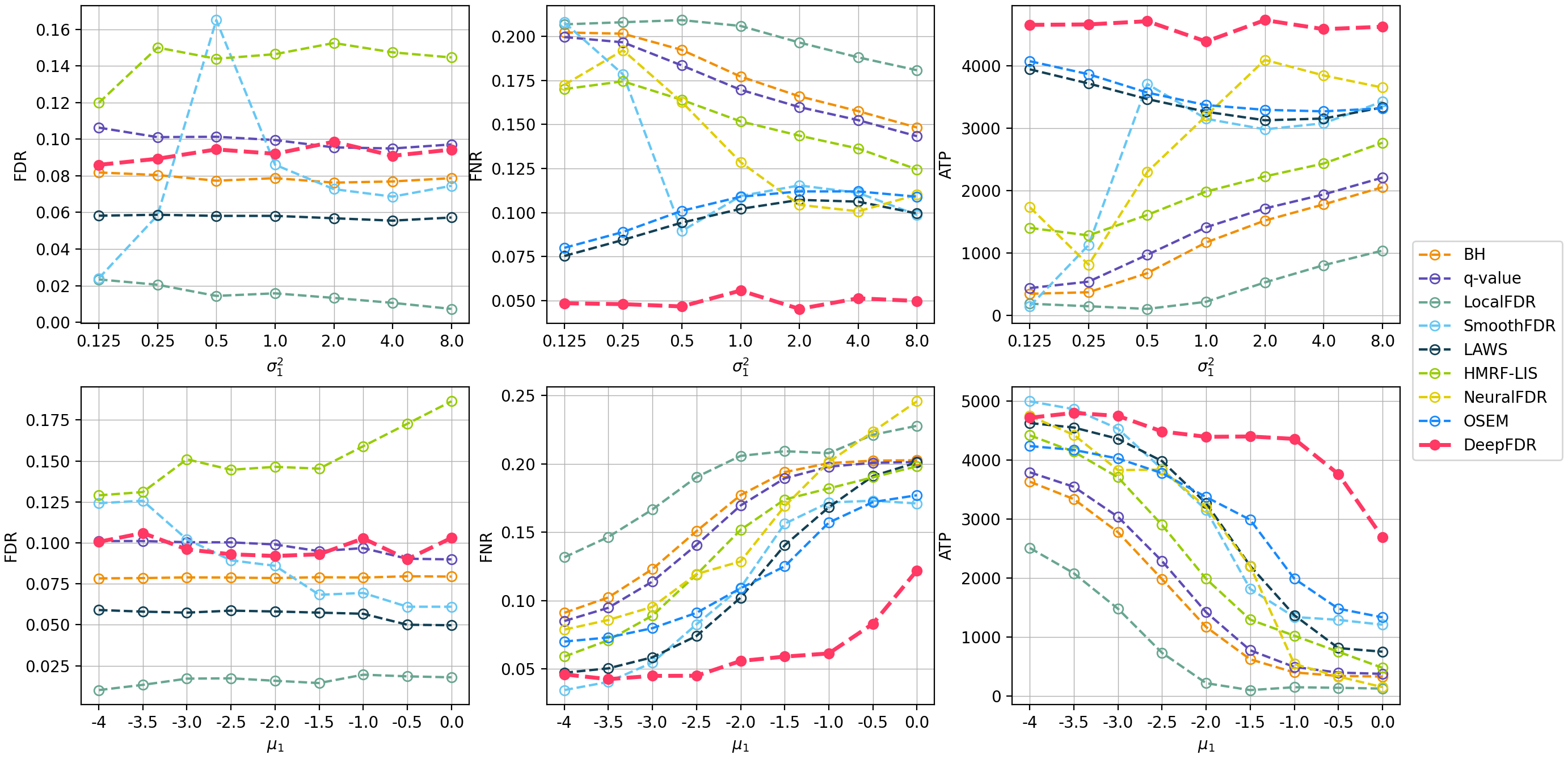}
\vspace{-0.4cm}
\caption{Simulation results for the cube with $P_1\approx 20\%$. FDRs for NeuralFDR and OSEM are too large and are thus not shown in this figure; see Figure~\ref{fig:simulation_0.2}, instead.}
\label{fig: siml p=0.2}
\end{figure*}

\begin{figure*}[h!] 
\centering
\includegraphics[width=0.85\textwidth]{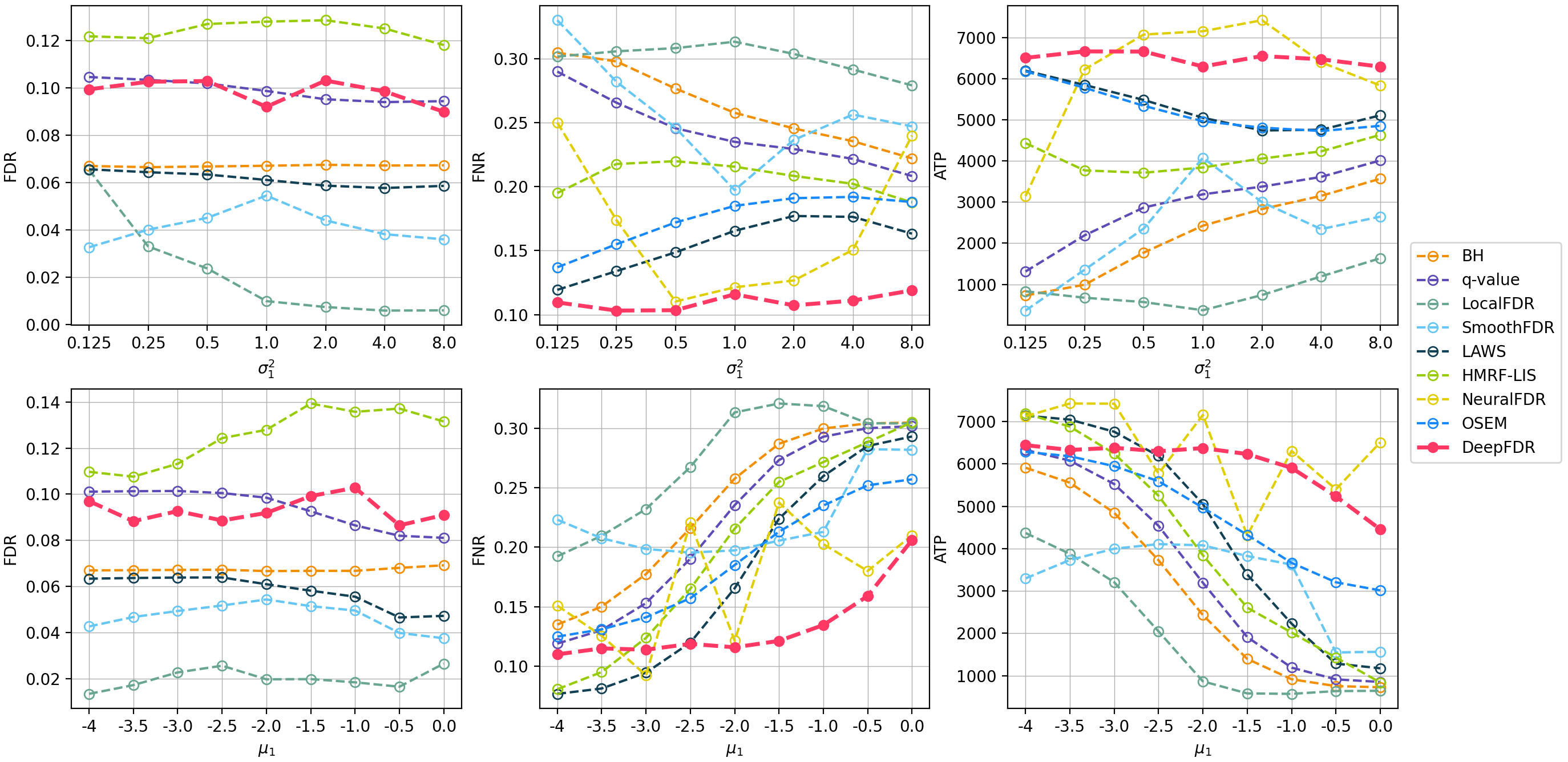}
\vspace{-0.4cm}
\caption{Simulation results for the cube with $P_1\approx 30\%$. FDRs for NeuralFDR and OSEM are too large and are thus not shown in this figure; see Figure~\ref{fig:simulation_0.3}, instead.}
\label{fig: siml p=0.3}
\end{figure*}

\begin{figure*}[h!] 
\centering
\includegraphics[width=\textwidth]{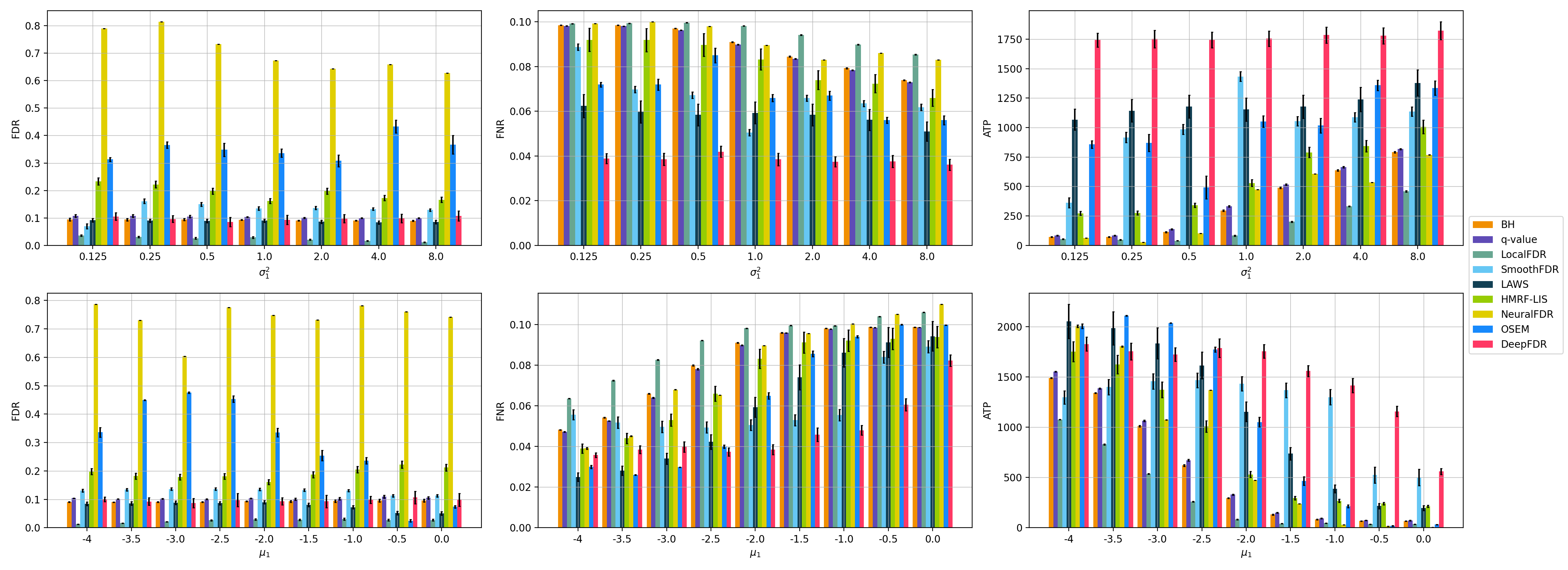}
\vspace{-1cm}
\caption{Simulation results with standard error bars for the cube with $P_1\approx 10\%$.}
\label{fig:simulation_0.1}
\end{figure*}

\begin{figure*}[h!] 
\centering
\includegraphics[width=\textwidth]{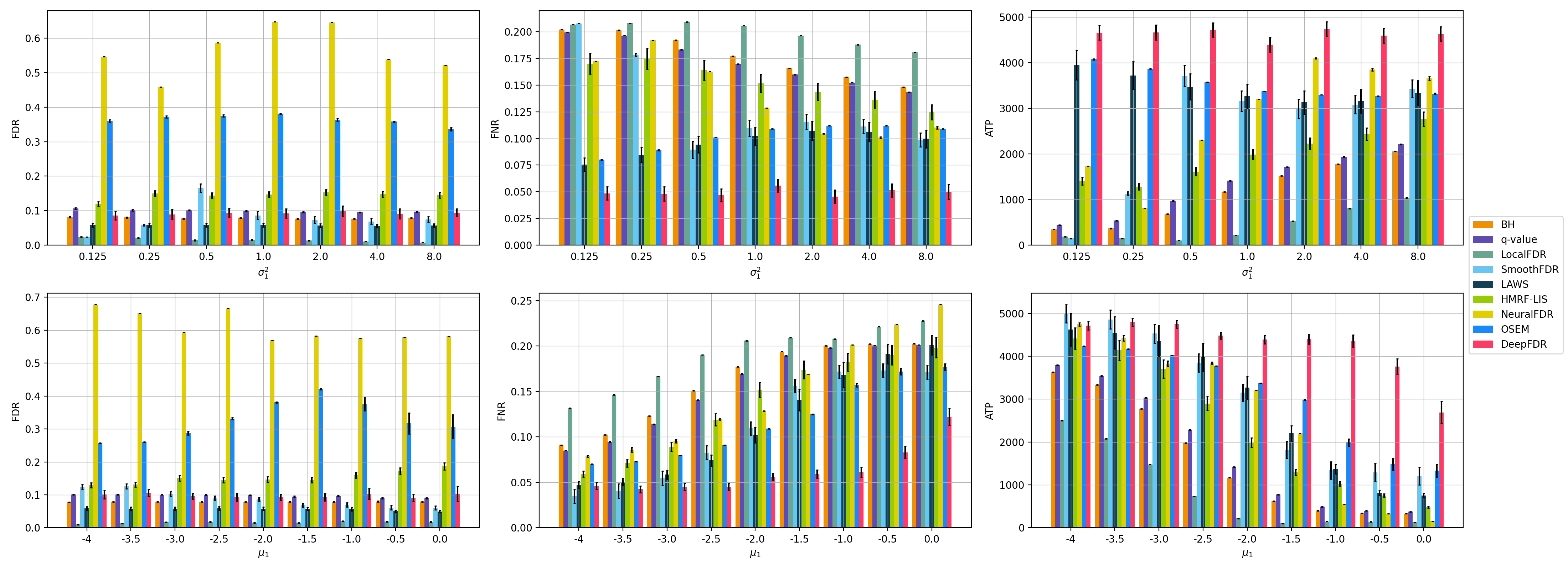}
\vspace{-1cm}
\caption{Simulation results with standard error bars for the cube with $P_1\approx 20\%$.}
\label{fig:simulation_0.2}
\end{figure*}

\begin{figure*}[h!] 
\centering
\includegraphics[width=\textwidth]{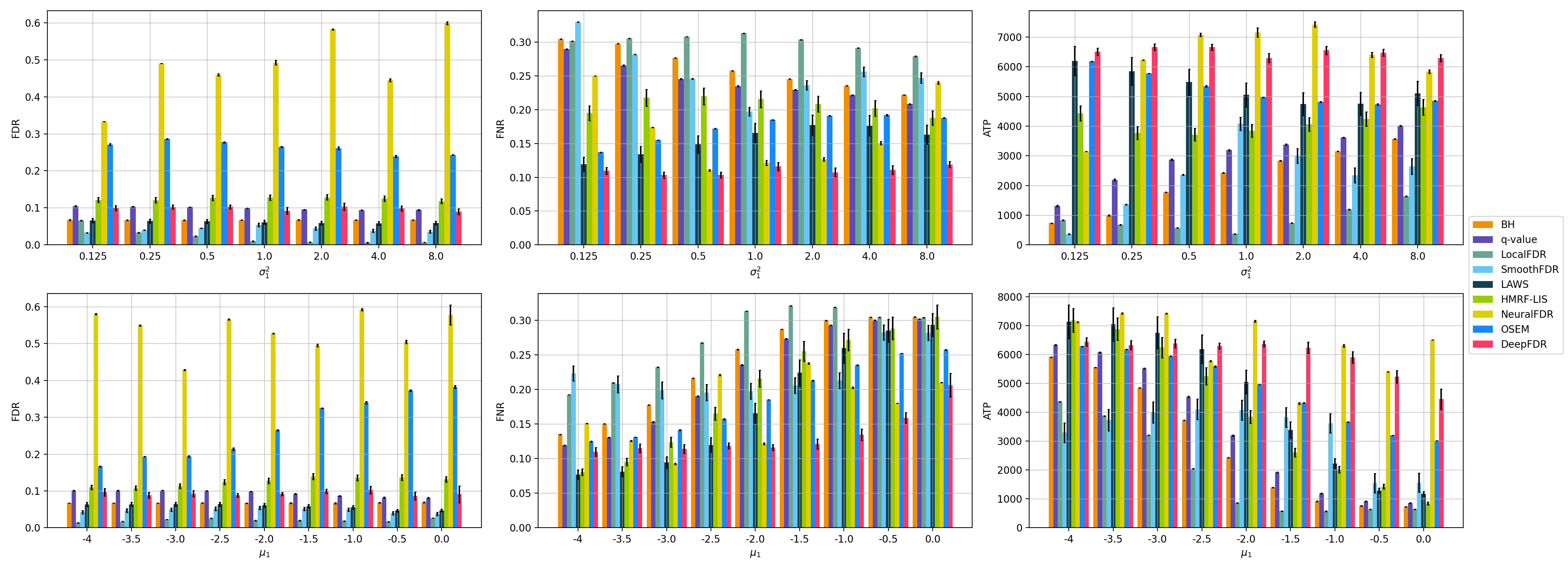}
\vspace{-1cm}
\caption{Simulation results with standard error bars for the cube with $P_1\approx 30\%$.}
\label{fig:simulation_0.3}
\end{figure*}

\begin{table}[h!]
\centering
\begin{tabular}{c|c|c}
\hline
    {Method} & {Simulation} & {ADNI data} \\ 
\hline
    {BH}  & \text{0.0794 (0.0088)} & \text{0.1320 (0.0266)} \\
    {q-value}  & \text{4.2547 (0.0422)} & \text{68.458 (0.3169)} \\
    {LocalFDR}  & \text{0.1969 (0.0119)}& \text{0.4005 (0.0726)}\\
    {SmoothFDR}  & \text{143.92 (2.0182)} & \text{53281 (3437.2)}\\
    {LAWS}  & \text{371.49 (1.2788)}& \text{611620 (24745)}\\
        {HMRF-LIS}  & \text{56.932 (6.2486)}& \text{20245 (1987.0)}\\
    {NeuralFDR} & \text{6205.1 (412.93)} & \text{95388 (6198.1)}\\
    {OSEM} & \text{15.565 (5.2412)} & \text{93312 (21603)}\\
        {DeepFDR}  & \text{7.2104 (1.2248)} & \text{89.984 (5.1672)} \\
\hline
\end{tabular}
\caption{Mean (and SD) of runtime in seconds.}
\label{table:run time for simulation}
\end{table}

\begin{figure*}[h!] 
\centering
\includegraphics[width=\textwidth]{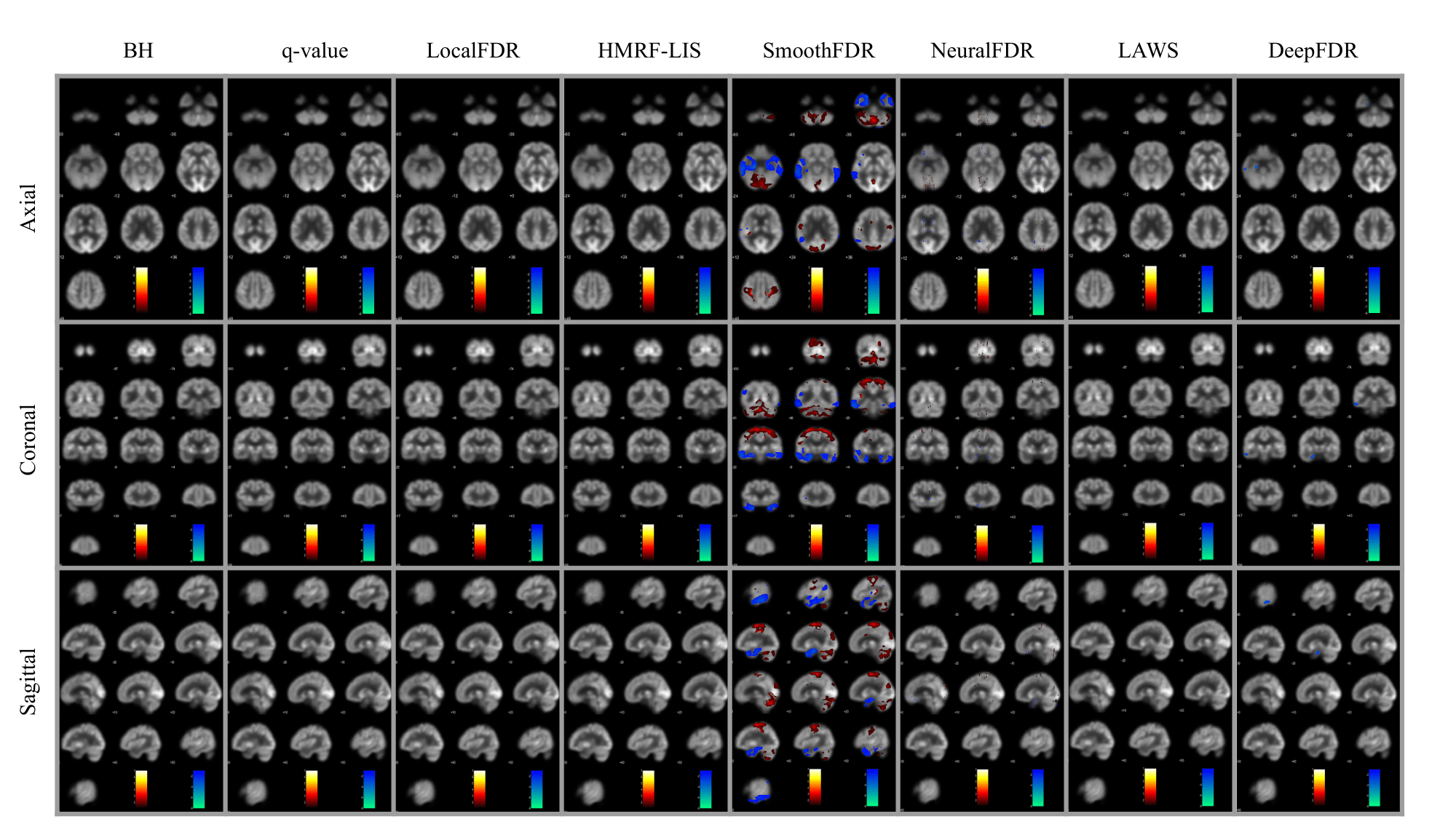}
\caption{Z-statistics of the discoveries by each considered method for EMCI2AD vs. CN. OSEM found no discoveries and is thus omitted.}
\label{fig: zvalue EMCI2AD}
\end{figure*}

\begin{figure*}[h!] 
\vspace{1cm}
\centering
\includegraphics[width=\textwidth]{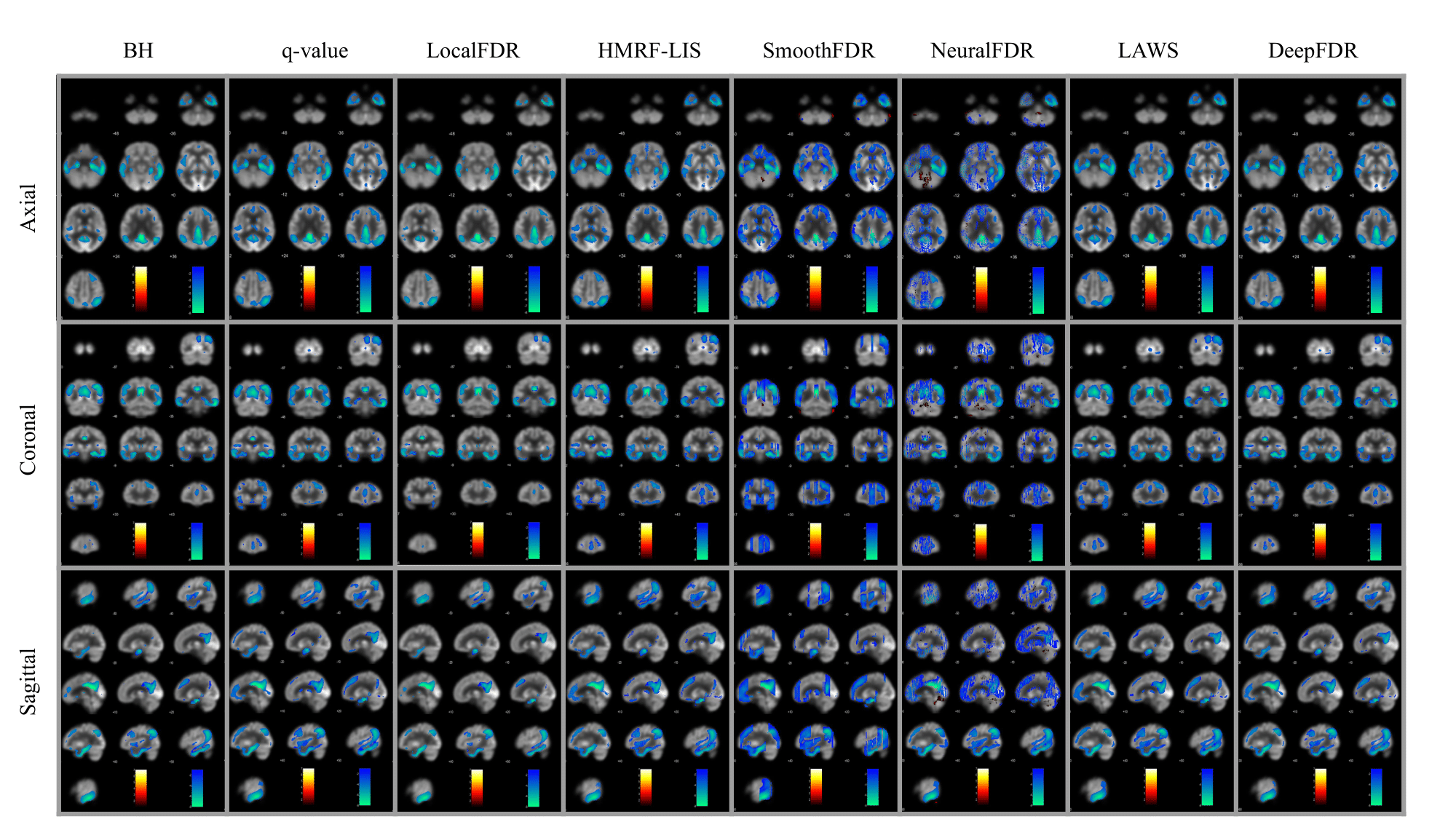}
\caption{Z-statistics of the discoveries by each considered method for LMCI2AD vs. CN. OSEM found no discoveries and is thus omitted.}
\label{fig: zvalue LMCI2AD}
\end{figure*}

\begin{figure*}[h!] 
\centering
\includegraphics[width=\textwidth]{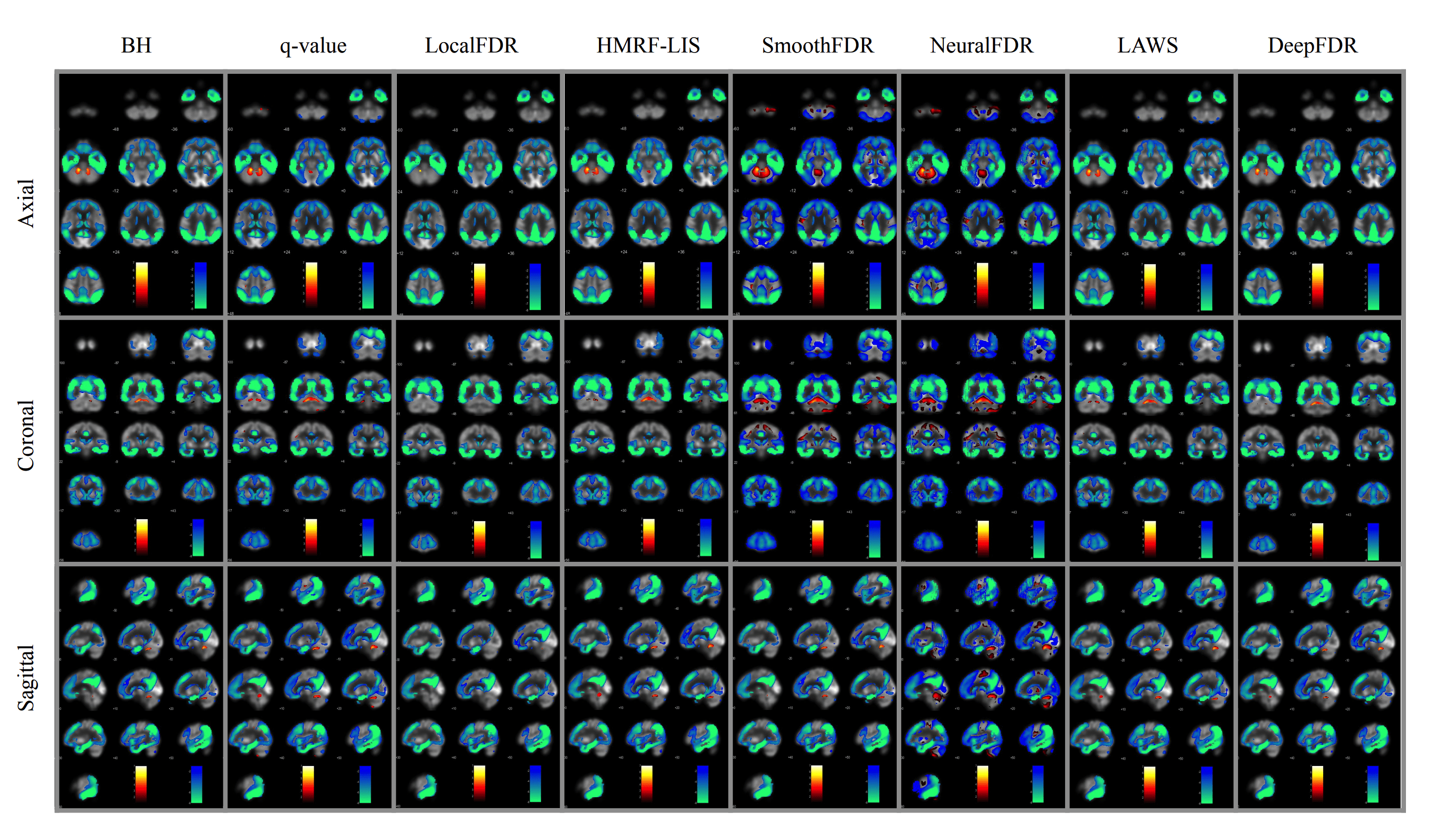}
\caption{Z-statistics of the discoveries by each considered method for AD vs. CN. OSEM found no discoveries and is thus omitted.}
\label{fig: zvalue AD}
\end{figure*}

\begin{figure*}[h!] 
\centering
\includegraphics[width=\textwidth]{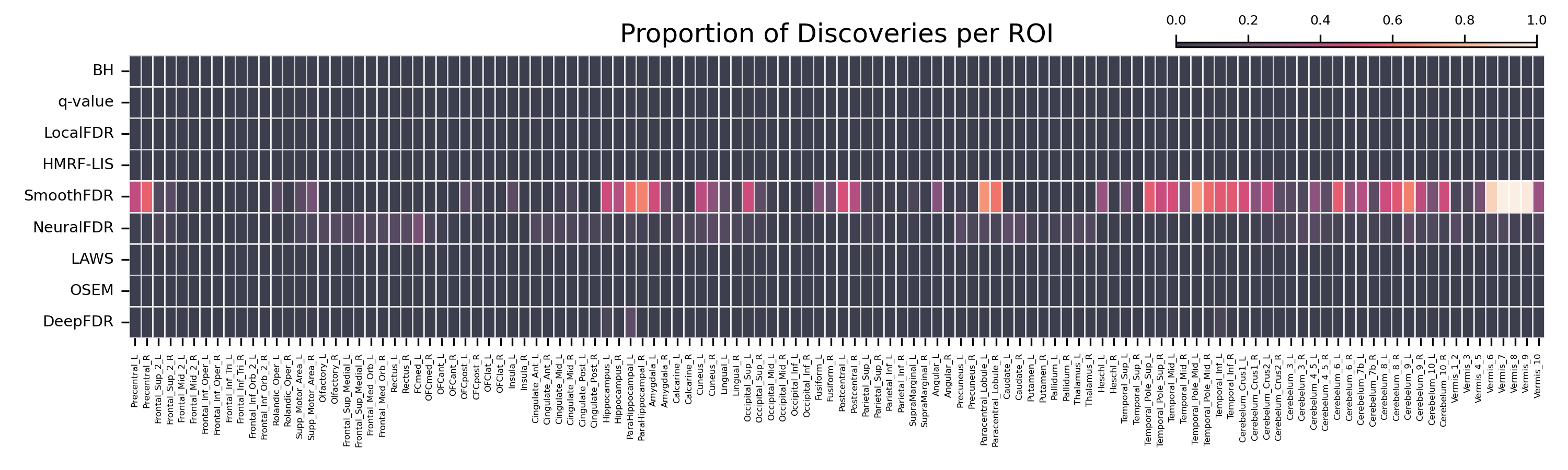}
\caption{Heatmap  illustrating  
the proportion of discoveries 
in each ROI for EMCI2AD vs. CN.}
\label{fig:HP EMCI}
\end{figure*}

\begin{figure*}[h!] 
\centering
\includegraphics[width=\textwidth]{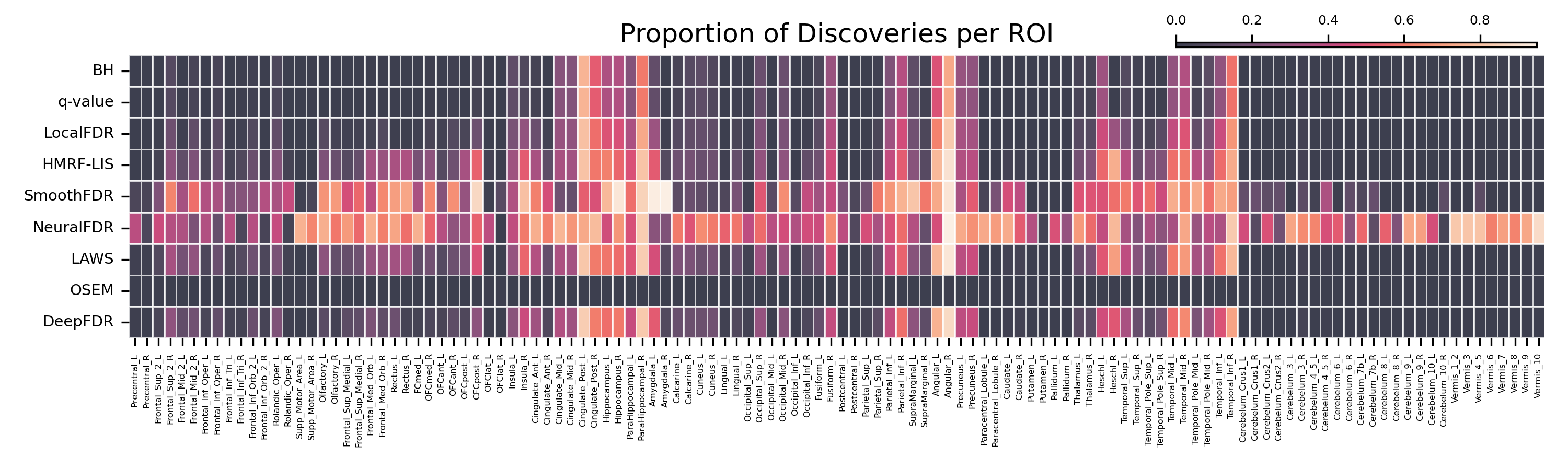}
\caption{Heatmap  illustrating  
the proportion of discoveries 
in each ROI for LMCI2AD vs. CN.}
\label{fig:HP LMCI}
\end{figure*}

\begin{figure*}[h!] 
\centering
\includegraphics[width=\textwidth]{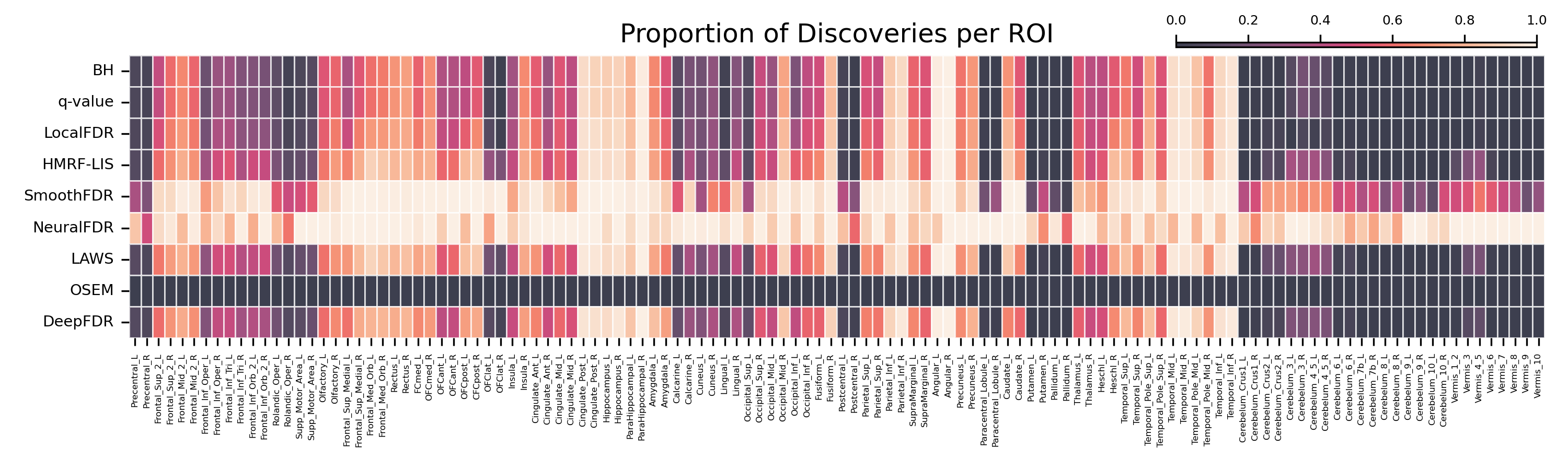}
\caption{Heatmap  illustrating  
the proportion of discoveries 
in each ROI for AD vs. CN.}
\label{fig:HP AD}
\end{figure*}

\begin{table*}[h!]
\centering
\begin{tabular}{cc|cc|cc}
\hline
\textbf{ROI} & \textbf{\# voxels} & \textbf{ROI} & \textbf{\# voxels} & \textbf{ROI} & \textbf{\# voxels} \\
\hline
Precentral\_L & 8281 & Precentral\_R & 7972 & Frontal\_Sup\_2\_L & 11315 \\
Frontal\_Sup\_2\_R & 12201 & Frontal\_Mid\_2\_L & 10701 & Frontal\_Mid\_2\_R & 11617 \\
Frontal\_Inf\_Oper\_L & 2496 & Frontal\_Inf\_Oper\_R & 3303 & Frontal\_Inf\_Tri\_L & 6020 \\
Frontal\_Inf\_Tri\_R & 5213 & Frontal\_Inf\_Orb\_2\_L & 1754 & Frontal\_Inf\_Orb\_2\_R & 1877 \\
Rolandic\_Oper\_L & 2405 & Rolandic\_Oper\_R & 3210 & Supp\_Motor\_Area\_L & 5057 \\
Supp\_Motor\_Area\_R & 5861 & Olfactory\_L & 648 & Olfactory\_R & 726 \\
Frontal\_Sup\_Medial\_L & 7178 & Frontal\_Sup\_Medial\_R & 4881 & Frontal\_Med\_Orb\_L & 1793 \\
Frontal\_Med\_Orb\_R & 2176 & Rectus\_L & 1950 & Rectus\_R & 1759 \\
FCmed\_L & 1272 & OFCmed\_R & 1457 & OFCant\_L & 1137 \\
OFCant\_R & 1631 & OFCpost\_L & 1410 & CFCpost\_R & 1401 \\
OFClat\_L & 488 & OFClat\_R & 475 & Insula\_L & 4418 \\
Insula\_R & 4204 & Cingulate\_Ant\_L & 3289 & Cingulate\_Ant\_R & 3230 \\
Cingulate\_Mid\_L & 4487 & Cingulate\_Mid\_R & 5169 & Cingulate\_Post\_L & 1079 \\
Cingulate\_Post\_R & 767 & Hippocampus\_L & 2225 & Hippocampus\_R & 2265 \\
ParaHippocampal\_L & 2330 & ParaHippocampal\_R & 2675 & Amygdala\_L & 504 \\
Amygdala\_R & 599 & Calcarine\_L & 5392 & Calcarine\_R & 4473 \\
Cuneus\_L & 3716 & Cuneus\_R & 3291 & Lingual\_L & 4945 \\
Lingual\_R & 5398 & Occipital\_Sup\_L & 3179 & Occipital\_Sup\_R & 3382 \\
Occipital\_Mid\_L & 7876 & Occipital\_Mid\_R & 4865 & Occipital\_Inf\_L & 2133 \\
Occipital\_Inf\_R & 2401 & Fusiform\_L & 5410 & Fusiform\_R & 5976 \\
Postcentral\_L & 9295 & Postcentral\_R & 9045 & Parietal\_Sup\_L & 4853 \\
Parietal\_Sup\_R & 5234 & Parietal\_Inf\_L & 5753 & Parietal\_Inf\_R & 3221 \\
SupraMarginal\_L & 2961 & SupraMarginal\_R & 4536 & Angular\_L & 2786 \\
Angular\_R & 4129 & Precuneus\_L & 8253 & Precuneus\_R & 7862 \\
Paracentral\_Lobule\_L & 3217 & Paracentral\_Lobule\_R & 2035 & Caudate\_L & 2280 \\
Caudate\_R & 2377 & Putamen\_L & 2392 & Putamen\_R & 2532 \\
Pallidum\_L & 665 & Pallidum\_R & 635 & Thalamus\_L & 2667 \\
Thalamus\_R & 2600 & Heschl\_L & 525 & Heschl\_R & 579 \\
Temporal\_Sup\_L & 5641 & Temporal\_Sup\_R & 7547 & Temporal\_Pole\_Sup\_L & 3005 \\
Temporal\_Pole\_Sup\_R & 3162 & Temporal\_Mid\_L & 11745 & Temporal\_Mid\_R & 10556 \\
Temporal\_Pole\_Mid\_L & 1789 & Temporal\_Pole\_Mid\_R & 2786 & Temporal\_Inf\_L & 7562 \\
Temporal\_Inf\_R & 8339 & Cerebelum\_Crus1\_L  & 6152 & Cerebelum\_Crus1\_R & 6258 \\
Cerebelum\_Crus2\_L & 4522 & Cerebelum\_Crus2\_R & 4994 & Cerebelum\_3\_L & 334 \\
Cerebelum\_3\_R & 536 & Cerebelum\_4\_5\_L & 2747 & Cerebelum\_4\_5\_R & 2086 \\
Cerebelum\_6\_L & 4113 & Cerebelum\_6\_R & 4291 & Cerebelum\_7b\_L & 1388 \\
Cerebelum\_7b\_R & 1276 & Cerebelum\_8\_L & 4454 & Cerebelum\_8\_R & 5490 \\
Cerebelum\_9\_L & 2069 & Cerebelum\_9\_R & 1956 & Cerebelum\_10\_L & 328 \\
Cerebelum\_10\_R & 374 & Vermis\_1\_2 & 107 & Vermis\_3 & 492 \\
Vermis\_4\_5 & 1442 & Vermis\_6 & 766 & Vermis\_7 & 468 \\
Vermis\_8 & 512 & Vermis\_9 & 412 & Vermis\_10 & 284 \\
\hline
\end{tabular}
\caption{The number of voxels in each ROI.}
\label{table:roi voxel number}
\end{table*}

\begin{table*}[h]
\centering
\begin{tabular}{ccccccccccc}
\hline
    {Method}  & {PHL} & {HL} & {TIL} & {TML} & {FL} & {TPML} & {TPSL} & {PREL} & {PRER} & {FS2L} \\ 
\hline
 {BH} & \text{0.0000} & \text{0.0000} & \text{0.0000} & \text{0.0000} & \text{0.0000} & \text{0.0000} & \text{0.0000} & \text{0.0000}  & \text{0.0000}  & \text{0.0000} \        \\
 {q-value} & \text{0.0000} & \text{0.0000} & \text{0.0000} & \text{0.0000} & \text{0.0000} & \text{0.0000} & \text{0.0000} & \text{0.0000}  & \text{0.0000}  & \text{0.0000} \   \\
 {LocalFDR} & \text{0.0000} & \text{0.0000} & \text{0.0000} & \text{0.0000} & \text{0.0000} & \text{0.0000} & \text{0.0000} & \text{0.0000}  & \text{0.0000}  & \text{0.0000} \  \\
 {HMRF-LIS} & \text{0.0000} & \text{0.0000} & \text{0.0000} & \text{0.0000} & \text{0.0000} & \text{0.0000} & \text{0.0000} & \text{0.0001}  & \text{0.0000}  & \text{0.0000} \  \\
 {SmoothFDR} & \text{0.6009} & \text{0.4710} & \text{0.5440} & \text{0.4928} & \text{0.2396} & \text{0.7289} & \text{0.5524} & \text{0.4268}  & \text{0.5696}  & \text{0.0860} \ \\
 {NeuralFDR} & \text{0.0476} & \text{0.0521} & \text{0.0057} & \text{0.0066} & \text{0.0043} & \text{0.0028} & \text{0.0027} & \text{0.0085}  & \text{0.0103}  & \text{0.0675} \ \\
 {LAWS} & \text{0.0000} & \text{0.0000} & \text{0.0000} & \text{0.0000} & \text{0.0000} & \text{0.0000} & \text{0.0000} & \text{0.0000}  & \text{0.0000}  & \text{0.0000} \      \\
  {OSEM} & \text{0.0000} & \text{0.0000} & \text{0.0000} & \text{0.0000} & \text{0.0000} & \text{0.0000} & \text{0.0000} & \text{0.0000}  & \text{0.0000}  & \text{0.0000} \      \\
 {DeepFDR} & \text{0.1185} & \text{0.0584} & \text{0.0518} & \text{0.0208} & \text{0.0070} & \text{0.0028} & \text{0.0007} & \text{0.0000}  & \text{0.0000}  & \text{0.0000} \   \\
\hline
\end{tabular}
\caption{Proportion of discoveries in the top 10 affected ROIs detected by DeepFDR for EMCI2AD vs. CN. See Table~\ref{table:roi_codes} for region abbreviations.}
\label{table:Results}
\end{table*}

\begin{table*}[h]
\centering
\begin{tabular}{ccccccccccc}
\hline
    {Method}  & {ANR} & {CPL} & {PHR} & {ANL} & {TIR} & {MTR} & {CPR} & {HR} & {PIR} & {HL} \\ 
\hline
 {BH} & \text{0.7266} & \text{0.7618} & \text{0.6064} & \text{0.4856} & \text{0.5864} & \text{0.3633} & \text{0.5254} & \text{0.3620}  & \text{0.3772}  & \text{0.3537} \        \\
 {q-value} & \text{0.7266} & \text{0.7618} & \text{0.6064} & \text{0.4856} & \text{0.5864} & \text{0.3633} & \text{0.5254} & \text{0.3620}  & \text{0.3772}  & \text{0.3537} \   \\
  {LocalFDR} & \text{0.8302} & \text{0.7998} & \text{0.7338} & \text{0.6242} & \text{0.6775} & \text{0.4941} & \text{0.5763} & \text{0.4773}  & \text{0.4617}  & \text{0.4921} \  \\
 {HMRF-LIS} & \text{0.9026} & \text{0.8054} & \text{0.8090} & \text{0.7757} & \text{0.7349} & \text{0.6106} & \text{0.5997} & \text{0.5545}  & \text{0.5253}  & \text{0.6225} \  \\
 {SmoothFDR} & \text{0.9121} & \text{0.5329} & \text{0.8561} & \text{0.7297} & \text{0.6998} & \text{0.6509} & \text{0.4811} & \text{0.9161}  & \text{0.7566}  & \text{0.7766} \ \\
 {NeuralFDR} & \text{0.9489} & \text{0.7294} & \text{0.8348} & \text{0.4648} & \text{0.8274} & \text{0.7230} & \text{0.7836} & \text{0.6777}  & \text{0.5709}  & \text{0.4512} \ \\
 {LAWS} & \text{0.9157} & \text{0.8174} & \text{0.8378} & \text{0.7721} & \text{0.7754} & \text{0.6852} & \text{0.6115} & \text{0.5744}  & \text{0.5473}  & \text{0.5960} \      \\
   {OSEM} & \text{0.0000} & \text{0.0000} & \text{0.0000} & \text{0.0000} & \text{0.0000} & \text{0.0000} & \text{0.0000} & \text{0.0000}  & \text{0.0000}  & \text{0.0000} \      \\
 {DeepFDR} & \text{0.8762} & \text{0.8378} & \text{0.8262} & \text{0.7538} & \text{0.7152} & \text{0.6438} & \text{0.6141} & \text{0.6031}  & \text{0.5815}  & \text{0.5748} \   \\
\hline
\end{tabular}
\caption{Proportion of discoveries in the top 10 affected ROIs detected by DeepFDR for LMCI2AD vs. CN. See Table~\ref{table:roi_codes} for region abbreviations.}
\label{table:Results}
\end{table*}

\begin{table*}[h]
\centering
\begin{tabular}{ccccccccccc}
\hline
    {Method}  & {ANR} & {PHR} & {AL} & {TMR} & {TIR} & {PIR} & {TML} & {HR} & {CPL} & {TIL} \\ 
\hline
 {BH} & \text{1.0000} & \text{0.9869} & \text{0.9871} & \text{0.9605} & \text{0.9621} & \text{0.9255} & \text{0.9367} & \text{0.8971}  & \text{0.9323}  & \text{0.9162} \        \\
 {q-value} & \text{1.0000} & \text{0.9869} & \text{0.9871} & \text{0.9605} & \text{0.9621} & \text{0.9255} & \text{0.9367} & \text{0.8971}  & \text{0.9323}  & \text{0.9162} \   \\
 {LocalFDR} & \text{1.0000} & \text{0.9918} & \text{0.9896} & \text{0.9737} & \text{0.9704} & \text{0.9419} & \text{0.9537} & \text{0.9227}  & \text{0.9527}  & \text{0.9312} \  \\
 {HMRF-LIS} & \text{1.0000} & \text{0.9940} & \text{0.9878} & \text{0.9798} & \text{0.9734} & \text{0.9497} & \text{0.9658} & \text{0.9426}  & \text{0.9425}  & \text{0.9378} \  \\
 {SmoothFDR} & \text{1.0000} & \text{1.0000} & \text{1.0000} & \text{1.0000} & \text{0.9999} & \text{1.0000} & \text{0.9999} & \text{1.0000}  & \text{1.0000}  & \text{0.9985} \ \\
 {NeuralFDR} & \text{1.0000} & \text{1.0000} & \text{0.8726} & \text{1.0000} & \text{1.0000} & \text{1.0000} & \text{0.8163} & \text{1.0000}  & \text{1.0000}  & \text{0.8360} \ \\
 {LAWS} & \text{1.0000} & \text{0.9963} & \text{0.9910} & \text{0.9673} & \text{0.9797} & \text{0.9581} & \text{0.9743} & \text{0.9435}  & \text{0.9731}  & \text{0.9501} \      \\
   {OSEM} & \text{0.0000} & \text{0.0000} & \text{0.0000} & \text{0.0000} & \text{0.0000} & \text{0.0000} & \text{0.0000} & \text{0.0000}  & \text{0.0000}  & \text{0.0000} \      \\
 {DeepFDR} & \text{1.0000} & \text{0.9981} & \text{0.9910} & \text{0.9821} & \text{0.9800} & \text{0.9733} & \text{0.9726} & \text{0.9660}  & \text{0.9592}  & \text{0.9528} \   \\
\hline
\end{tabular}
\caption{Proportion of discoveries in the top 10 affected ROIs detected by DeepFDR for AD vs. CN. See Table~\ref{table:roi_codes} for region abbreviations.}
\label{table:Results}
\end{table*}

\begin{table*}[h!]
\centering
\begin{tabular}{cc|cc|cc}\hline
 \textbf{Code} &  \textbf{ROI Name} &  \textbf{Code} &  \textbf{ROI Name} &  \textbf{Code} &  \textbf{ROI Name} \\
\hline
PREL & Precentral\_L & PRER & Precentral\_R & FS2L & Frontal\_Sup\_2\_L \\
FS2R & Frontal\_Sup\_2\_R & FM2L & Frontal\_Mid\_2\_L & FM2R & Frontal\_Mid\_2\_R \\
FIOL & Frontal\_Inf\_Oper\_L & FIOR & Frontal\_Inf\_Oper\_R & FITL & Frontal\_Inf\_Tri\_L \\
FITR & Frontal\_Inf\_Tri\_R & FIO2L & Frontal\_Inf\_Orb\_2\_L & FIO2R & Frontal\_Inf\_Orb\_2\_R \\
ROL & Rolandic\_Oper\_L & ROR & Rolandic\_Oper\_R & SMAL & Supp\_Motor\_Area\_L \\
SMAR & Supp\_Motor\_Area\_R & OL & Olfactory\_L & OR & Olfactory\_R \\
FSML & Frontal\_Sup\_Medial\_L & FSMR & Frontal\_Sup\_Medial\_R & FMOL & Frontal\_Med\_Orb\_L \\
FMOR & Frontal\_Med\_Orb\_R & RL & Rectus\_L & RR & Rectus\_R \\
FCL & FCmed\_L & OFCMR & OFCmed\_R & OFCAL & OFCant\_L \\
OFCAR & OFCant\_R & OFCPL & OFCpost\_L & CFCR & CFCpost\_R \\
OFCLL & OFClat\_L & OFCLR & OFClat\_R & IL & Insula\_L \\
IR & Insula\_R & CAL & Cingulate\_Ant\_L & CAR & Cingulate\_Ant\_R \\
CML & Cingulate\_Mid\_L & CMR & Cingulate\_Mid\_R & CPL & Cingulate\_Post\_L \\
CPR & Cingulate\_Post\_R & HL & Hippocampus\_L & HR & Hippocampus\_R \\
PHL & ParaHippocampal\_L & PHR & ParaHippocampal\_R & AL & Amygdala\_L \\
AR & Amygdala\_R & CAL & Calcarine\_L & CAR & Calcarine\_R \\
CL & Cuneus\_L & CR & Cuneus\_R & LL & Lingual\_L \\
LR & Lingual\_R & OSL & Occipital\_Sup\_L & OSR & Occipital\_Sup\_R \\
OML & Occipital\_Mid\_L & OMR & Occipital\_Mid\_R & OIL & Occipital\_Inf\_L \\
OIR & Occipital\_Inf\_R & FL & Fusiform\_L & FR & Fusiform\_R \\
POSTL & Postcentral\_L & POSTR & Postcentral\_R & PSL & Parietal\_Sup\_L \\
PSR & Parietal\_Sup\_R & PIL & Parietal\_Inf\_L & PIR & Parietal\_Inf\_R \\
SML & SupraMarginal\_L & SMR & SupraMarginal\_R & ANL & Angular\_L \\
ANR & Angular\_R & PCL & Precuneus\_L & PCR & Precuneus\_R \\
PLL & Paracentral\_Lobule\_L & PLR & Paracentral\_Lobule\_R & CAUL & Caudate\_L \\
CAUR & Caudate\_R & PUL & Putamen\_L & PUR & Putamen\_R \\
PAL & Pallidum\_L & PAR & Pallidum\_R & TL & Thalamus\_L \\
TR & Thalamus\_R & HEL & Heschl\_L & HER & Heschl\_R \\
TSL & Temporal\_Sup\_L & TSR & Temporal\_Sup\_R & TPSL & Temporal\_Pole\_Sup\_L \\
TPSR & Temporal\_Pole\_Sup\_R & TML & Temporal\_Mid\_L & TMR & Temporal\_Mid\_R \\
TPML & Temporal\_Pole\_Mid\_L & TPMR & Temporal\_Pole\_Mid\_R & TIL & Temporal\_Inf\_L \\
TIR & Temporal\_Inf\_R & CC1L & Cerebelum\_Crus1\_L  & CC1R & Cerebelum\_Crus1\_R \\
CC2L & Cerebelum\_Crus2\_L & CC2R & Cerebelum\_Crus2\_R & C3L & Cerebelum\_3\_L \\
C3R & Cerebelum\_3\_R & C45L & Cerebelum\_4\_5\_L & C45R & Cerebelum\_4\_5\_R \\
C6L & Cerebelum\_6\_L & C6R & Cerebelum\_6\_R & C7L & Cerebelum\_7b\_L \\
C7R & Cerebelum\_7b\_R & C8L & Cerebelum\_8\_L & C8R & Cerebelum\_8\_R \\
C9L & Cerebelum\_9\_L & C9R & Cerebelum\_9\_R & C10L & Cerebelum\_10\_L \\
C10R & Cerebelum\_10\_R & V12 & Vermis\_1\_2 & V3 & Vermis\_3 \\
V45 & Vermis\_4\_5 & V6 & Vermis\_6 & V7 & Vermis\_7 \\
V8 & Vermis\_8 & V9 & Vermis\_9 & V10 & Vermis\_10 \\
\hline
\end{tabular}
\caption{Abbreviation codes for ROI names.}
\label{table:roi_codes}
\end{table*}

\end{document}